\documentclass[sigconf,screen]{acmart}

\usepackage{fancyhdr}
\usepackage{graphicx}
\usepackage[dvipsnames,svgnames,table,xcdraw]{xcolor}
\usepackage{threeparttable}
\usepackage{amsmath}
\usepackage{url}
\usepackage{multirow}
\usepackage{booktabs}
\usepackage{enumitem}
\usepackage[most]{tcolorbox}
\usepackage{enumitem}
\usepackage{float} 
\usepackage{fvextra}
\usepackage{adjustbox}
\usepackage{colortbl}
\usepackage[normalem]{ulem}
\usepackage{xr-hyper}
\usepackage{pifont}
\usepackage{tabularx}
\usepackage{placeins}

\definecolor{lightbluegray}{RGB}{220, 230, 242}
\definecolor{lightyellow}{RGB}{255, 243, 210}
\definecolor{lightmint}{RGB}{220, 238, 222}
\definecolor{lightrosegray}{RGB}{240, 220, 220}
\definecolor{lightlavender}{RGB}{235, 230, 245}
\definecolor{lightgray}{RGB}{235,235,235}

\renewcommand\footnotetextcopyrightpermission[1]{}

\AtBeginDocument{%
  }

\setcopyright{acmlicensed}
\copyrightyear{2018}
\acmYear{2018}
\acmDOI{XXXXXXX.XXXXXXX}

\acmConference[Preprint]{arXiv}{Under Review}{April 2026}

\acmISBN{978-1-4503-XXXX-X/2018/06}
\acmSubmissionID{529}

\usepackage{etoolbox}
\makeatletter
\patchcmd{\authornote}{\g@addto@macro\addresses{\@authornotemark}}{}{}{}
\makeatother
\begin{document}

\title{AeSlides: Incentivizing Aesthetic Layout in LLM-Based Slide Generation via Verifiable Rewards}

\author{%
Yiming Pan$^{1}$\footnote[1]{} \quad
Chengwei Hu$^{2}$ \quad
Xuancheng Huang$^{2}$\footnote[2]{} \quad
Can Huang$^{2}$ \quad
Mingming Zhao$^{2}$ \quad \\
Yuean Bi$^{2}$ \quad
Xiaohan Zhang$^{2}$ \quad
Aohan Zeng$^{2}$ \quad
Linmei Hu$^{1}$\footnote[2]{}%
}
\authornote{ Work done during internship at Zhipu AI.}
\authornote{ Corresponding authors.}

\affiliation{%
  \institution{$^{1}$ Beijing Institute of Technology \qquad $^{2}$ Zhipu AI (Z.ai)}
  \country{}
}

\renewcommand{\shortauthors}{Pan et al.}

\begin{abstract}
Large language models (LLMs) have demonstrated strong potential in agentic tasks, particularly in slide generation. However, slide generation poses a fundamental challenge: the generation process is text-centric, whereas its quality is governed by visual aesthetics.
This modality gap leads current models to frequently produce slides with aesthetically suboptimal layouts. Existing solutions typically rely either on heavy visual reflection, which incurs high inference cost yet yields limited gains; or on fine-tuning with large-scale datasets, which still provides weak and indirect aesthetic supervision.
In contrast, the explicit use of aesthetic principles as supervision remains unexplored.
In this work, we present \textbf{AeSlides}, a reinforcement learning framework with verifiable rewards for \textbf{Ae}sthetic layout supervision in \textbf{Slide} generation. 
We introduce a suite of meticulously designed verifiable metrics to quantify slide layout quality, capturing key layout issues in an accurate, efficient, and low-cost manner.
Leveraging these verifiable metrics, we develop a GRPO-based reinforcement learning method that directly optimizes slide generation models for aesthetically coherent layouts.
With only 5K training prompts on GLM-4.7-Flash, AeSlides improves aspect ratio compliance from 36\% to 85\%, while reducing whitespace by 44\%, element collisions by 43\%, and visual imbalance by 28\%. Human evaluation further shows a substantial improvement in overall quality, increasing scores from 3.31 to 3.56 (+7.6\%), outperforming both model-based reward optimization and reflection-based agentic approaches, and even edging out Claude-Sonnet-4.5. 
These results demonstrate that such a verifiable aesthetic paradigm provides an efficient and scalable approach to aligning slide generation with human aesthetic preferences.
Our code, datasets, and model checkpoints are publicly available at \url{https://github.com/ympan0508/aeslides}.
\end{abstract}



\begin{teaserfigure}
    \adjustbox{valign=c,vspace=2pt}{
        \includegraphics[width=\textwidth]{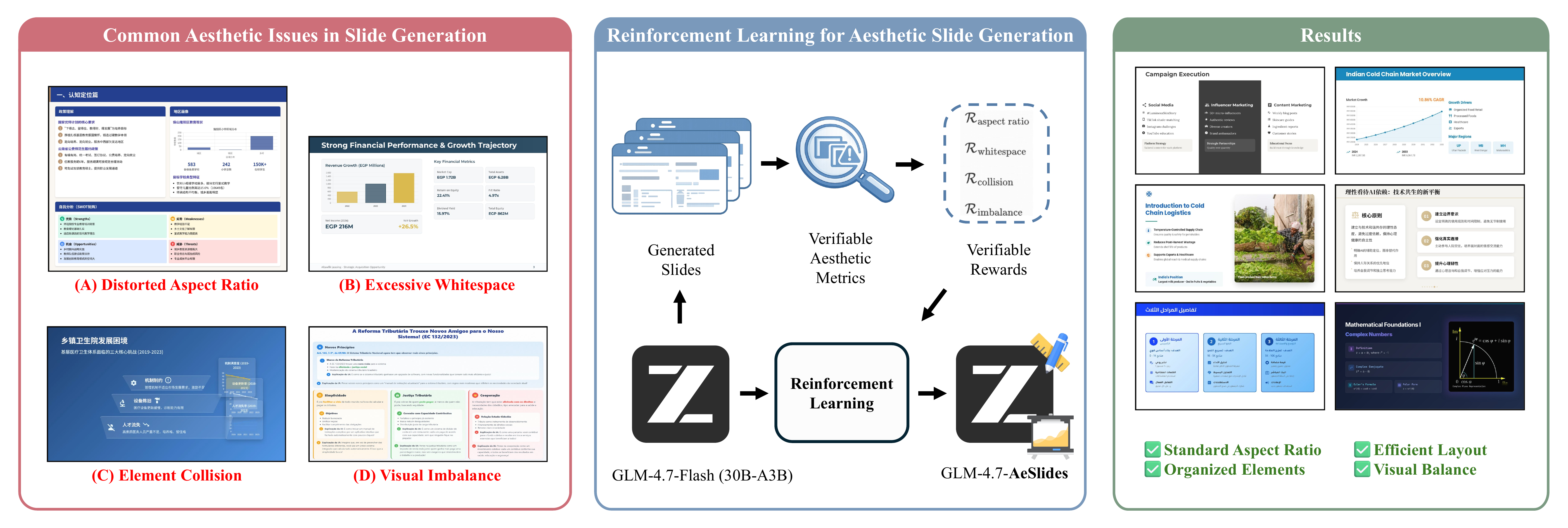}
    }
    \caption{Overview of the AeSlides workflow. \textit{Left}: Four categories of aesthetic deficiencies commonly observed in LLM-based slide generation. \textit{Center}: A suite of verifiable aesthetic metrics is introduced and integrated into reinforcement learning to guide the model toward producing visually coherent slide layouts. \textit{Right}: Representative slides generated by GLM-4.7-AeSlides.}
  \label{fig:teaser}
\end{teaserfigure}

\maketitle
\section{Introduction}


Slides are a widely used medium for communicating structured information in domains such as education, business, and scientific presentation. 
While recent advances in agentic capabilities of large language models (LLMs) have enabled substantial progress in complex tasks such as software engineering~\cite{Yang2024SWEagentAI}, slide generation remains particularly challenging, as it requires not only semantic coherence but also precise control over visual layout.
In practice, current models often fail to produce aesthetically sound layouts. While LLMs typically generate slides as structured \textit{textual} markup, slide quality is inherently evaluated in the \textit{visual} domain. This \textit{modality gap} between text-based generation and vision-based evaluation limits the effectiveness of text-only supervision, leading even proprietary models such as Claude-Sonnet-4.5 and GPT-5.2 to frequently produce slides with aesthetic layout issues.

Early slide generation approaches primarily rely on predefined templates~\cite{fu2022doc2ppt, cachola2024knowledge}, which restrict flexibility and fail to accommodate diverse user requirements. Recent works~\cite{zheng2025pptagent,liang2025slidegen, tang2025slidecoder, ge2025autopresent, xupregenie,zheng2026deeppresenter} adopt agentic frameworks that perform iterative refinement, leveraging multimodal feedback to identify and correct rendering defects. While effective for coarse issues such as severely broken layouts, these methods are constrained by the limited visual perception capabilities and often fail to address fine-grained aesthetic issues. Another line of work~\cite{zeng2025glm} attempts to fine-tune LLMs on large-scale curated datasets to implicitly learn aesthetic preferences; however, the modality gap continues to hinder further improvement.
This leads to a key question: \textit{Can aesthetic competence be directly internalized into the model, rather than corrected through post-generation refinement or implicit supervision?}
We observe that many fundamental aesthetic properties of slide layouts are inherently structured and can be precisely verified through programmatic analysis. This observation suggests an alternative paradigm: aesthetic  layout quality can be explicitly optimized through verifiable reward signals.

To address these challenges, we propose \textbf{AeSlides}, a reinforcement learning framework with verifiable rewards for improving aesthetic slide generation. We first construct two datasets targeting slide aesthetic quality: one for meta-evaluation of aesthetic metrics and another for reinforcement learning. Based on four key aesthetic issues observed in slide generation (as illustrated in Figure~\ref{fig:teaser}), we introduce a suite of meticulously designed verifiable metrics. Meta-evaluation results show that, compared to vision-language models (VLM)-based detections, these metrics achieve higher accuracy, lower latency, and reduced cost across all dimensions. Leveraging these metrics, we develop a Group Relative Policy Optimization~\cite{shao2024deepseekmath} (GRPO)-based reinforcement learning method that directly optimizes aesthetic quality via verifiable reward signals.

Extensive experiments demonstrate that AeSlides substantially improves aesthetic layout quality in slide generation. On GLM-4.7-Flash, using only 5k training prompts, AeSlides increases the aspect ratio compliance rate from 36\% to 85\%, reduces excessive whitespace by 44\%, decreases element collisions by 43\%, and mitigates visual imbalance by 28\%. Human evaluation further confirms a significant improvement in overall quality, raising the score from 3.31 to 3.56. 
Additionally, with single-pass generation, AeSlides consistently outperforms reflection-based iterative methods, surpasses model-based reward optimization, and even edges out proprietary Claude-Sonnet-4.5.
These results validate the effectiveness of reinforcement learning with verifiable aesthetic rewards for slide generation. Further analysis also reveals the unreliability of model-based metrics and rewards, highlighting that our verifiable reward design is a key factor in improving slide layout aesthetics.

In summary, our contributions are as follows:

\begin{enumerate}[leftmargin=*, topsep=0pt, partopsep=0pt]
    \item We introduce a reinforcement learning with verifiable rewards paradigm in slide generation, enabling direct optimization of layout aesthetics to mitigate the modality gap.
    \item We design a suite of accurate and efficient verifiable metrics that decompose layout aesthetics into measurable subproblems, supporting both RL optimization and reliable slide evaluation.
    \item Extensive experiments demonstrate consistent improvements across all metrics and overall evaluation, outperforming strong baselines and validating the effectiveness of verifiable rewards.
\end{enumerate}

\section{Related Work}

\subsection{LLM-Based Slide Generation}

Benefiting from capabilities in language understanding and generation, LLMs have become the primary backbone of slide generation. Early approaches mainly use LLMs for content generation, while relying on predefined templates for layout and visual design, as in DOC2PPT~\cite{fu2022doc2ppt} and KCTV~\cite{cachola2024knowledge}. However, template-based methods struggle to accommodate diverse requirements in layout flexibility, visual styling, and compositional design. With the rise of agentic LLMs, recent systems decompose slide generation into structured workflows involving background research, global planning, and tool-based editing, such as PPTAgent~\cite{zheng2025pptagent}, AutoPresent~\cite{ge2025autopresent}, and Talk-to-Your-Slides~\cite{jung2025talk}. Despite improved flexibility, these methods still lack explicit aesthetic constraints. As a result, they either produce over-conservative layouts (e.g., simple bullet-point lists) or exhibit unstable visual quality when handling complex designs.

Recent work attempts to address the modality gap between text-based generation and vision-based evaluation along two directions. One adopts post-generation visual reflection for iterative refinement~\cite{zheng2026deeppresenter, liang2025slidegen, xupregenie}, but incurs high inference cost while yielding only limited improvement in aesthetic quality. The other direction relies on supervised training over large-scale curated slide datasets~\cite{zeng2025glm}, where aesthetic signals are still implicitly learned via textual patterns, resulting in indirect and weak supervision.
In contrast, directly internalizing aesthetic principles into the base model through training remains largely underexplored. 
As advances in coding capabilities~\cite{zeng2026glm, team2026kimi} enable LLMs to directly generate structured slide representations in a single pass (e.g., via \texttt{HTML} or \texttt{Slidev}), the lack of explicit aesthetic modeling becomes more critical.

\subsection{Reinforcement Learning in LLMs}
In-domain adaptation of LLMs to acquire specialized expertise has been extensively studied. Early approaches primarily rely on SFT to inject domain knowledge and capabilities, while subsequent methods leverage human feedback, either through direct preference alignment (DPO)~\cite{rafailov2023direct} or online reinforcement learning (RL)~\cite{ouyang2022training}. 
However, the offline nature of DPO and the computational burden of online RLHF have limited their scalability across diverse vertical domains. 
The introduction of GRPO~\cite{shao2024deepseekmath} provides a lightweight yet effective alternative, enabling arbitrary reward signals to be directly optimized. This paradigm has demonstrated strong performance in domains with verifiable feedback, such as coding~\cite{deepcoder2025} and mathematics~\cite{shao2024deepseekmath}. Follow-up works~\cite{yu2025dapo,zheng2025gspo, ma2025stabilizing, gao2025soft, IcePop2025, yao2025offpolicy, liu2026gdpo} further improve training stability and effectiveness, especially for MoE models, long-context tasks, and multi-reward settings, consolidating GRPO as a practical choice for reinforcement learning.

However, in slide generation, applying RL remains challenging due to the abstract nature of aesthetic requirements. In related domains such as webpage and UI generation, prior work~\cite{jiangwebgen, yang2025ui, Patnaik2025AesthetiQEG} employs VLMs as reward models or rubric generators to provide reward signals. In contrast, our experiments reveal that current VLMs exhibit notable aesthetic blind spots for slide layouts, which limits the transferability of such approaches to slide generation. In this work, we instead explore an alternative paradigm by introducing a suite of meticulously designed verifiable metrics that decompose aesthetic layout issues into verifiable subproblems commonly encountered in slide generation. These metrics are used as vision-based rewards to train slide generation models. This approach provides a practical way to narrow the modality gap between generation and evaluation, enabling models to produce slides with improved aesthetic quality aligned with human preferences.

\section{Task Definition}
\label{sec:task-def}

Slide generation is inherently an agentic task. In its general form, it constitutes a long-horizon process: given a user query $Q$, a model $G_\theta$ first needs to acquire a richer supporting context $C$ through steps such as external information retrieval, image search, user intent clarification, and high-level planning. Conditioned on $Q$ and $C$, the model then sequentially generates a slide deck in HTML format, producing slides page by page until completion.
In this work, to enable clear disentanglement and attribution of the model's capability in generating aesthetically coherent slides, we focus on a controlled subproblem: \emph{prefix-conditioned next-slide generation}. Specifically, at step $t$, given the query $Q$, the context $C$, and the preceding slides $P_{<t}$, the model generates the next slide
\begin{equation}
\label{eq:task-def}
P_t \sim G_\theta(Q, C, P_{<t}).
\end{equation}
We then evaluate and optimize $P_t$ with respect to aesthetic metrics.

Concretely, current slide generation models often exhibit several recurring issues in layout aesthetics, which can be mainly categorized into four types:
(i) \emph{distorted aspect ratio}, where the model fails to account for the spatial extent of generated content, leading to layouts that deviate significantly from the standard $16{:}9$ aspect ratio; 
(ii) \emph{excessive whitespace}, where the page is not effectively utilized, resulting in large unused regions and a visually sparse appearance; 
(iii) \emph{element collision}, where elements overlap with each other, exceed their parent container, or fall outside the slide boundary; 
and (iv) \emph{visual imbalance}, where poor layout leads to uneven content distribution and a perceptual bias of visual weight toward one side of the slide.
We illustrate one representative example of each category in Figure~\ref{fig:teaser}. In this work, we focus on addressing these four issues as primary factors affecting aesthetic slide quality.

\section{Data Preparation}

We first collect a set of slide deck generation trajectories produced by LLMs, covering multiple languages and domains. Using rule-based preprocessing, we filter out failed trajectories and decompose the remaining ones into page-level samples, corresponding to the prefix-conditioned formulation defined in Equation~\ref{eq:task-def}. On top of this processed data, we construct two datasets to support aesthetic evaluation and optimization of slide generation:

\textbf{(i) AeSlides-Reward-Bench.} This dataset is designed for meta-evaluation of aesthetic metrics. We recruit and train six annotators with prior experience in slide design. They annotate slide pages with a fine-grained taxonomy of aesthetic flaws, consisting of 4 major categories and 21 subcategories, including the primary layout-related issues addressed in this work (as described in Section~\ref{sec:task-def}), while the remaining subcategories are reserved for future exploration. We publicly release this dataset to facilitate further research on additional optimization dimensions for slide generation.

\textbf{(ii) AeSlides-7k.} This dataset is used for training and evaluating slide generation models. We apply heuristic rules to remove structurally simple pages, including cover pages, tables of contents, dividers, and ending pages. These pages are excluded because existing models already achieve relatively strong aesthetic quality on them, and the design principles they follow differ from those of content-intensive slides. The resulting dataset is then split at the deck level into training and evaluation sets with a ratio of 85:15.

Detailed annotation protocols, metadata and statistics of the datasets are provided in Appendix~\ref{appx:data-meta} in supplementary materials.

\section{Methodology}
\label{sec:method}

\subsection{Verifiable Metrics for Slide Assessment}
\label{sec:metrics}

\begin{table*}[t]
\centering
\small
\begin{threeparttable}
\scalebox{0.899}{
\begin{tabular}{l|ccc|ccc|ccc|ccc|c|c}
\toprule
\multirow{2}{*}{Metric Type} 
& \multicolumn{3}{c|}{Distorted Aspect Ratio} 
& \multicolumn{3}{c|}{Excessive Whitespace} 
& \multicolumn{3}{c|}{Element Collision} 
& \multicolumn{3}{c|}{Visual Imbalance}  
& \multirow{2}{*}{Latency $\downarrow$}
& \multirow{2}{*}{Cost $\downarrow$}\\
& F1 $\uparrow$ & F2 $\uparrow$ & ROC-AUC $\uparrow$ 
& F1 $\uparrow$ & F2 $\uparrow$ & ROC-AUC $\uparrow$ 
& F1 $\uparrow$ & F2 $\uparrow$ & ROC-AUC $\uparrow$ 
& F1 $\uparrow$ & F2 $\uparrow$ & ROC-AUC $\uparrow$ 
& & \\ 
\midrule

\rowcolor{lightgray} Coin-Flip Baseline
& 0.58 & 0.54 & 0.50
& 0.51 & 0.50 & 0.50
& 0.13 & 0.23 & 0.50
& 0.22 & 0.33 & 0.50
& -
& - \\

GPT-5.2 
& 0.24 & 0.17 & 0.56
& 0.73 & 0.82 & 0.71
& 0.43 & 0.45 & 0.70
& 0.36 & 0.57 & 0.77
& $\sim$8000ms
& $\sim$\$200\\

GPT-5-mini 
& 0.54 & 0.43 & 0.68
& 0.41 & 0.31 & 0.61
& 0.19 & 0.13 & 0.55
& 0.47 & 0.56 & 0.75
& $\sim$10000ms
& $\sim$\$30\\

GPT-5-nano 
& 0.34 & 0.25 & 0.58
& 0.59 & 0.59 & 0.58
& 0.18 & 0.14 & 0.55
& 0.42 & 0.56 & 0.75
& $\sim$12000ms
& $\sim$\$6\\

Verifiable Metrics (Ours) 
& \textbf{0.83} & \textbf{0.91} & \textbf{0.77} 
& \textbf{0.80} & \textbf{0.88} & \textbf{0.89}
& \textbf{0.47} & \textbf{0.51} & \textbf{0.79} 
& \textbf{0.61} & \textbf{0.77} & \textbf{0.83}  
& \textbf{$\sim$4000ms}
& \textbf{$\sim$\$0} \\

\bottomrule
\end{tabular}
}
\end{threeparttable}
\caption{\label{tab:meta-eval} Meta-evaluation of our verifiable aesthetic layout metrics and VLM-based detection. \textit{Cost} denotes the estimated detection cost per 50k samples; detailed estimation method is provided in Appendix~\ref{appx:cost}. \textit{Latency} refers to the end-to-end detection latency, both including approximately $\sim$3000ms rendering time. For VLM-based methods, latency further includes network request overhead and remote model inference time, while for verifiable methods, it further includes metric computation time. }
\end{table*}

To address the four aesthetic issues defined in Section~\ref{sec:task-def}, we design a suite of rule-based, verifiable metrics for automatic detection.

\textbf{Render Infrastructure}. 
We first build a distributed and decoupled rendering server based on Playwright, FastAPI, and Uvicorn to support high-throughput rendering and metric computation during rollout and evaluation. Upon receiving HTML content, the server renders the page while injecting JavaScript scripts to collect runtime metadata and properties (e.g., page sizes, document object model (DOM) trees, and element bounding-boxes). The rendered image bytes, original HTML, and runtime properties are then passed to corresponding metric functions as needed. The entire pipeline is executed in a streaming and concurrent manner, yielding high throughput. Decoupling rollout from metric computation also allows training hardware to focus on rollout generation and parameter updates, thereby improving overall training efficiency.

\textbf{Detecting distorted aspect ratio}. 
Since most slides adopt responsive layouts, measuring the actual aspect ratio requires a dynamic procedure. We first initialize a small viewport (1280 $\times$ 10) to load the page. After rendering is complete, a JavaScript script is executed to retrieve the maximum value among the \texttt{scrollHeight} and \texttt{offsetHeight} of both \texttt{body} and \texttt{html}. The viewport height is then reset to this value, and a screenshot is taken to obtain the final width and height of the rendered slide. 
 
\begin{figure}[h]
    \centering
    \includegraphics[width=\linewidth]{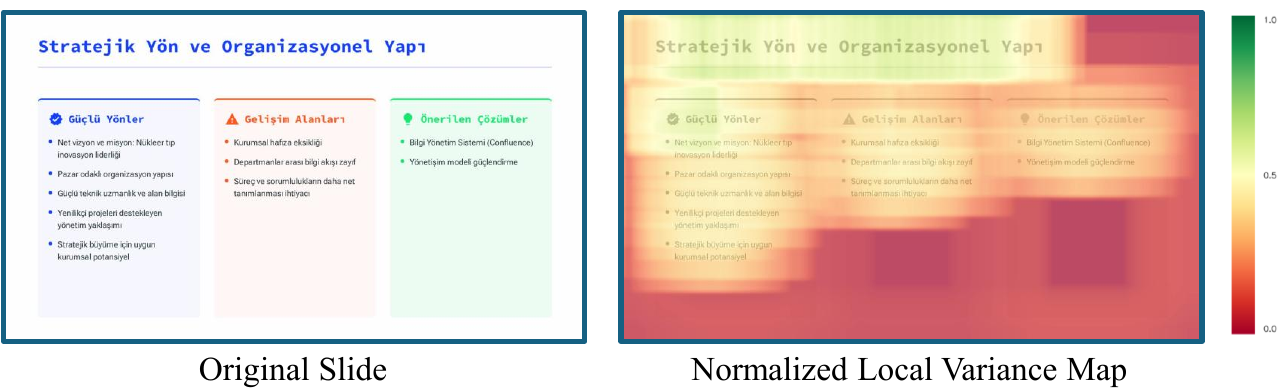}
    \caption{Visualization of our excessive whitespace detection. \textit{Left}: original slide. \textit{Right}: the slide overlaid with a normalized local variance map, where red regions indicate low local pixel variation and are identified as whitespace.}
    \label{fig:whitespace}
\end{figure}

\textbf{Detecting excessive whitespace}. 
We adopt a lightweight vision-based method for whitespace detection. Specifically, we define ``whitespace'' as regions lacking significant local pixel variation, and compute a local variance map to estimate the proportion of effective content. Specifically, the procedure consists of: (i) convert the image $I$ to grayscale, yielding $I'$; 
(ii) use box filters to compute the local variance map $\mathrm{Var}(x,y)$ over a rectangular neighborhood $\Omega \subset \mathbb{Z}^2$ centered at $(x,y)$ with size $H \times W$ (we use $H{=}201, W{=}151$):
\begin{equation}
\mu(x,y) = \frac{1}{|\Omega|} \sum_{(i,j)\in\Omega} I'(x+i, y+j)
\end{equation}
\begin{equation}
\mathrm{Var}(x,y) = \frac{1}{|\Omega|} \sum_{(i,j)\in\Omega} \left(I'(x+i, y+j) - \mu(x,y)\right)^2
\end{equation}
(iii) obtain the standard deviation $\sigma(x,y)=\sqrt{\mathrm{Var}(x,y)}$, followed by clipping and normalization to calculate the local variance map
\begin{equation}
F(x,y) = \frac{\min(\sigma(x,y), T_{\text{clip}})}{T_{\text{clip}}}
\end{equation}
where smaller $F(x,y)$ indicates low-frequency regions corresponding to whitespace; 
(iv) binarizing the map using a threshold $\tau=0.05$, removing boundary regions (as peripheral redundancy is acceptable), and computing the ratio of content and whitespace areas. 
We visualize this process in Figure~\ref{fig:whitespace}, where red regions denote detected whitespace. The use of a large neighborhood implicitly provides tolerance for spacing between elements, avoiding overly fine-grained responses to intra-element gaps (e.g., character/line spacing). This makes the metric monotonic and easier to optimize.

\textbf{Detecting element collision}. 
We adopt heuristic rules based on the DOM tree and element bounding boxes. First, we extract ``visual units'' by filtering out invisible elements, non-semantic nodes, and overly fine-grained primitives (e.g., SVG primitives), to better align with human perception. Based on these units, we identify three types of collision events: (i) overlapping bounding boxes between elements without ancestor relationships; (ii) elements escaping from their parent containers; (iii) elements overflowing beyond the slide boundary. Certain cases that do not affect perceived quality (e.g., watermarks, weak transparent backgrounds, corner badges, annotation labels) are excluded unless they occlude main content. The remaining events are reweighted to compute a collision score, where higher scores indicate more severe layout conflicts.

\textbf{Detecting visual imbalance}. 
Similar to collision detection, visual imbalance is computed based on the DOM tree. After extracting visual units, we compute the visual centroid of the entire layout and measure its offset from the canvas center using an ellipse-normalized distance. Let $(x,y)\in[0,1]^2$ denote the normalized coordinates of the visual centroid, and $(x_c, y_c)$ the canvas center. The ellipse-normalized distance is defined as
\begin{equation}
d = \sqrt{\left(\frac{x - x_c}{x_{\text{tol}}}\right)^2 + \left(\frac{y - y_c}{y_{\text{tol}}}\right)^2},
\end{equation}
where $x_{\text{tol}}$ and $y_{\text{tol}}$ are tolerance parameters along two axes. Since horizontal imbalance is more perceptually salient, we set $x_{\text{tol}}=0.05$ and $y_{\text{tol}}=0.15$. Larger values of $d$ indicate more severe imbalance.

\textbf{Reward Hacking Mitigation}. 
During training, we observe that the model may exploit these metrics via reward hacking. Instead of introducing explicit penalties (which empirically lead to rapid abandonment of error-prone patterns and collapse of the policy space), we iteratively refine the rendering pipelines to eliminate exploitable shortcuts. Specifically: 
(i) for aspect ratio, the model tends to enforce layout constraints using attributes such as \texttt{min-height} and \texttt{overflow: hidden}, which deviates from our objective of learning proper content quantity and layout adjustments. We therefore normalize the HTML using \texttt{cssutils} and remove such attributes in the aspect ratio pipeline;
(ii) for whitespace detection, the model may insert textured backgrounds or external background images to artificially increase local variance. To address this, we remove background images and apply a small Gaussian filter (kernel size $=21$) before box filtering to smooth texture-induced variance.

\textbf{Meta-evaluation and comparison with VLM-based detection}. 
To validate the effectiveness of the proposed metrics, we additionally integrate VLM-based detection as a metric within the render infrastructure. Given a rendered slide, the VLM outputs a score in the range of 0--5 for each of the four issue dimensions, where lower scores indicate more severe problems. 
We conduct a meta-evaluation on the four aesthetic issue categories using AeSlides-Reward-Bench, comparing the proposed verifiable metrics with VLM-based detection. We report F1, F2, ROC-AUC, the average end-to-end latency, and the cost per 50k samples (comparable to the number required to train one epoch on our dataset). The classification threshold is selected according to the optimal F2 criterion. The latency includes image rendering time ($\sim$3000ms). The corresponding detection prompts are provided in Appendix~\ref{appx:vlm-issue-detection}.
As shown in Table~\ref{tab:meta-eval}, VLM-based methods perform poorly on most dimensions, in some cases even worse than random guessing. Notably, even a frontier model such as GPT-5.2 exhibits a clear failure in aspect ratio perception: despite being provided with explicit criteria, it tends to classify nearly all input slides as satisfying the aspect ratio requirement, even when the ratio is close to 1:1. 
In addition, VLM-based detection is less efficient, and also incurs higher cost. We further explore several variants, including enabling chain-of-thought reasoning~\cite{wei2022chain}, replacing scalar scores with binary outputs, and adopting G-Eval-style logit weighting~\cite{liu2023geval}; none of these approaches lead to improved detection accuracy. 
For Element Collision and Visual Imbalance, there exists an inherent ambiguity in the labels, where some samples are borderline cases, which leads to moderate F1 and F2 scores for our method. However, the consistently higher ROC-AUC indicates strong ranking ability, suggesting that the metrics remain suitable for reward computation in reinforcement learning.

\subsection{RL with Verifiable Rewards}

After constructing the metrics for evaluating the aesthetic quality of slide layouts, we employ them as reward signals in Reinforcement Learning (RL) to incentivize the model toward aesthetic layout in slide generation. We adopt Group Relative Policy Optimization (GRPO)~\cite{shao2024deepseekmath} as the starting point, which can be formulated as
\begin{equation}
\scalebox{0.83}{$
\begin{aligned}
\mathcal{J}_{\text{GRPO}}&(\theta)
= \mathbb{E}_{x_i\sim\mathcal{D}, \{y_j\}_{j=1}^{G} \sim \pi_{\theta_{\text{old}}}(\cdot|x_i)} 
\Bigg[\frac{1}{G} \sum_{j=1}^{G} \frac{1}{|y_j|} \sum_{t=1}^{|y_j|} \\
&\bigg(
\min\Big(
w_{j,t}(\theta)\hat{A}_{j,t},
\text{clip}\big(
w_{j,t}(\theta),1-\varepsilon,1+\varepsilon
\big)\hat{A}_{j,t}
\Big) 
-\beta D_{\text{KL}}(\pi_\theta\|\pi_{\text{ref}})
\bigg)
\Bigg]
\end{aligned}
$}
\end{equation}
Building upon this framework, we incorporate a set of auxiliary mechanisms that have been widely validated in prior work, including: (i) clip-higher and token-level loss proposed in DAPO~\cite{yu2025dapo}; (ii) sequence-level importance ratio introduced in GSPO~\cite{zheng2025gspo}; and (iii) rollout routing replay~\cite{ma2025stabilizing}. Prior work has both empirically validated and theoretically characterized these mechanisms, showing that they effectively stabilize training in Mixture-of-Experts LLMs as well as in long-context generation tasks.
In addition, we introduce several task-specific designs tailored to slide generation:

(i) \textbf{KL divergence regularization}. Recent studies~\cite{yu2025dapo, liuunderstanding} suggest that, in long-context generation tasks, the optimized policy may significantly deviate from the initialization, and therefore advocate removing the KL divergence constraint. In contrast, we retain the KL regularization term. Our base model has been supervised on a large corpus of high-quality slide data, ensuring strong content fidelity while exhibiting deficiencies in visual aesthetics. Our objective is to refine the model’s aesthetic behavior, which typically involves adjusting a subset of layout-related HTML attributes rather than substantially altering semantic content. Therefore, we do not expect significant distributional drift in model parameters. The KL term is thus preserved as a prior constraint to maintain alignment with the initialization model.

(ii) \textbf{Reward shaping}. The reinforcement learning setup in this work involves multiple reward components. Prior work~\cite{liu2026gdpo}
has shown that, under standard GRPO, different reward components may exhibit significantly different variance scales, which can lead to reward signal collapse during optimization. We also provide a simple theoretical justification in Appendix~\ref{appx:reward-collapse}. To address this issue, we design tailored reward shaping strategies.

For the aspect ratio metric, which follows a ``nominal-the-best'' pattern with an optimal value of $16{:}9$, the reward should decrease on both sides of the optimum. Empirically, we observe that overlong slides (i.e., smaller aspect ratios) occur more frequently, and thus decide to impose additional penalties on this regime. Specifically, we design an asymmetric quadratic reward
\begin{equation}
\mathcal{R}(x)=\exp\bigg(\!
-\alpha\big(\!
\log(\frac{x}{\text{target}})
\big)^2\!-\!\beta\cdot\max\big(\!-\!\log(\frac{x}{\text{target}})-m, 0 \big)^2
\bigg)
\end{equation}
where $x$ is the aspect ratio, \textit{target} is $16{:}9$, $m$ denotes the margin controlling tolerance for overlong cases, $\alpha$ adjusts the overall penalty, while $\beta$ controls the additional penalty for overlong deviations.

For the remaining three monotonic metrics, we adopt a unified strategy. We define \textit{lower} and \textit{upper} thresholds, beyond which the metric values are clipped to binary rewards (0 or 1), treating them as equally poor or equally good. Within this interval, we apply a \textit{smoothstep} transformation. Overall, the reward is formulated as
\begin{equation}
\scalebox{0.99}{$
\mathcal{R}(x) = \begin{cases}
1, &x\le \text{lower}\\
0, &x\ge \text{upper}\\
3u^2-2u^3, &\text{otherwise}
\end{cases}\quad \text{where} \; u=\frac{\text{upper} - x}{\text{upper}-\text{lower}}
$}
\end{equation}
where $x$ is the original metric value (higher indicates worse quality), \textit{lower} and \textit{upper} determine the truncation bounds.

(iii) \textbf{Reward-decoupled normalization}. 
To further stabilize multi-reward optimization, beyond reward shaping, we incorporate reward-decoupled normalization following GDPO~\cite{liu2026gdpo}. Specifically, when computing the advantage for rollout samples, each reward component is normalized independently. Let $r^{(k)}_{i,j}$ denote the $k$-th reward component for the $j$-th rollout under the $i$-th prompt
\begin{equation}
A_{i,j}^{(k)} = \frac{r_{i,j}^{(k)} - \mathbb{E}_{j' \sim \{1,\ldots,G\}}[r_{i,j'}^{(k)}]}{\mathrm{std}_{j' \sim \{1,\ldots,G\}} [r_{i,j'}^{(k)}] + \varepsilon}, \quad k=1,\ldots,K
\end{equation}
\begin{equation}
\quad A_{i,j} = \sum_k A_{i,j}^{(k)}
\end{equation}
where $G$ is the number of rollouts per prompt, and $K$ is the number of reward components.
Following~\cite{liu2026gdpo}, to maintain a stable numerical range, a batch-wise normalization is further applied
\begin{equation}
\hat{A}_{i,j} = \frac{A_{i,j} - \mathbb{E}_{i' \in D_{\text{batch}},\, j' \in \{1,\ldots,G\}} [A_{i',j'}]}{\mathrm{std}_{i' \in D_{\text{batch}},\, j' \in \{1,\ldots,G\}} [A_{i',j'}] + \varepsilon}
\end{equation}
The resulting advantage is then assigned to each rollout and broadcast to all tokens in the generated response.

(iv) \textbf{Error handling}. 
If no valid HTML is generated, or if the generated HTML fails to render, the overall reward is set to 0. Equivalently, this can be viewed as defining the overall reward as a conditional reward, which implicitly enforces output validity as the highest-priority requirement.

In summary, our policy optimization objective is formulated as
\begin{equation}
\scalebox{0.83}{$
\begin{aligned}
\mathcal{J}_\text{AeSlides}&(\theta)
= \mathbb{E}_{x_i\sim\mathcal{D}, \{y_j\}_{j=1}^{G} \sim \pi_{\theta_{\text{old}}}(\cdot|x_i)} 
\Bigg[\frac{1}{\textcolor{DarkOrange}{\sum_{j=1}^G |y_j|}} \textcolor{DarkOrange}{\sum_{j=1}^{G} \sum_{t=1}^{|y_j|}} \\
&\bigg(
\min\Big(
\textcolor{OliveGreen}{s_{j}(\theta)}\textcolor{SteelBlue}{\hat{A}_{i,j,t}},
\text{clip}\big(
\textcolor{OliveGreen}{s_{j}(\theta)},1-\textcolor{BrickRed}{\varepsilon_\text{low}},1+\textcolor{BrickRed}{\varepsilon_\text{high}}
\big)\textcolor{SteelBlue}{\hat{A}_{i,j,t}}
\Big) 
-\textcolor{Plum}{\beta D_{\text{KL}}(\pi_\theta\|\pi_{\text{ref}})}
\bigg)
\Bigg]
\end{aligned}
$}
\end{equation}
where the highlighted parts correspond to \textcolor{BrickRed}{clip-higher}, \textcolor{DarkOrange}{token-level loss}, \textcolor{OliveGreen}{sequence-level importance ratio}, \textcolor{SteelBlue}{reward-decoupled normalization}, \textcolor{Plum}{KL divergence}, respectively.

\section{Experiments}

\begin{table*}[!htbp]
\centering
\small
\begin{threeparttable}
\begin{tabular}{l|c|cccc|cc|c}
\toprule
                                   & \multicolumn{1}{c|}{}                               & \multicolumn{4}{c|}{\textbf{Verifiable Metrics}} & \multicolumn{2}{c|}{\textbf{VLM Score}} &                                    \\
\multirow{-2}{*}{\textbf{Model}}            & \multirow{-2}{*}{\textbf{Render Error} ↓} & A.R. ↑     & E.W. ↓    & E.C. ↓    & V.I. ↓    & GPT-5-mini ↑      & GPT-5.2 ↑     & \multirow{-2}{*}{\textbf{Human Evaluation} ↑} \\ \hline
\multicolumn{9}{l}{\textit{Proprietary Models}}                                                                                                                                                       \\ \hline
\cellcolor{lightrosegray}
GPT-5.2                            &         1.65\%                                           &  10\%/26\%        & 0.039        &    0.254     &  1.322       &        4.454         &  4.154           &     3.076                               \\
\cellcolor{lightrosegray}
Claude-Sonnet-4.5                  &      0.00\%                                              &   14\%/32\%       &     0.034    &  0.097       &  1.319       &      4.538           &  4.121           &      3.442                              \\ \hline
\multicolumn{9}{l}{\textit{Methodological Variants}}                                                                                                                                                  \\ \hline
\cellcolor{lightyellow}
Base Model (GLM-4.7-Flash) &      0.78\%                                              &    24\%/36\%      & 0.059        & 0.126        & 1.545        & 4.654                & 3.988            &    3.314                                \\
\cellcolor{lightmint}
+ DeepPresenter w/ heavy reflection                   &    1.74\%                                                &  27\%/41\%        & 0.070        &   0.212      &    2.170     &  4.396               &       3.875      &        3.107                            \\
\cellcolor{lightbluegray}
+ Static GPT-5-nano Reward         &     1.07\%                                               &  39\%/55\%        & 0.084        & 0.146        &     1.921            & 4.117            & 3.920                    & -              \\
\cellcolor{lightbluegray}
+ Static GPT-5-mini Reward         &    0.87\%                                                &   51\%/64\%       & 0.053        & 0.161        & 1.754        & 4.638                & 4.018            &      3.308                              \\
\cellcolor{lightbluegray}
+ Visual GPT-5-nano Reward         &     0.00\%                                               &   25\%/37\%       & 0.100        & 0.167        & 1.914        & 4.234                & 3.932            & -                                   \\
\cellcolor{lightbluegray}
+ Visual GPT-5-mini Reward         &    0.00\%                                                &  28\%/45\%        & 0.049        & 0.106        & 1.472        & \underline{4.754}                & 4.033            &             3.354                       \\
\cellcolor{lightlavender}
+ Verifiable Rewards               &       0.00\%                                             &  \underline{64\%}/\underline{75\%}        & \textbf{0.032}        & \underline{0.087}        & \textbf{1.071}        & 4.729                & \underline{4.042}            &       \underline{3.435}                             \\
\cellcolor{lightlavender}
+ Verifiable Rewards w/ GDPO        &        0.00\%                                            & \textbf{76\%}/\textbf{85\%}         & \underline{0.033}        & \textbf{0.072}        & \underline{1.111}        &      \textbf{4.771}           & \textbf{4.058}            &                 \textbf{3.561}                  \\
\bottomrule
\end{tabular}
\end{threeparttable}
\caption{\label{tab:main-results}Main experimental results. Different colors are used to highlight different types of variants: {\setlength{\fboxsep}{1pt}\colorbox{lightrosegray}{proprietary models}}, {\setlength{\fboxsep}{1pt}\colorbox{lightyellow}{base model}}, {\setlength{\fboxsep}{1pt}\colorbox{lightmint}{agentic reflection}}, {\setlength{\fboxsep}{1pt}\colorbox{lightbluegray}{RL with model-based reward}}, and {\setlength{\fboxsep}{1pt}\colorbox{lightlavender}{RL with verifiable rewards (\textit{Ours})}}. Within the \textit{Methodological Variants} group, the best results are highlighted in bold, and the second-best results are marked with underlining. A.R.: aspect ratio compliance under tolerances of 1\% and 5\%, E.W.: excessive whitespace, E.C.: element collision, V.I.: visual imbalance.}
\end{table*}

\subsection{Experimental Setup}

\textbf{Training.} We implement the method introduced in Section~\ref{sec:method} within the \texttt{slime}~\cite{slime_github} framework and perform RL training on GLM-4.7-Flash (30B-A3B)~\cite{zeng2025glm}. The model has been extensively trained on slide data during SFT; our goal is to further enhance its capability in generating aesthetically coherent slides. All experiments are conducted on a single node with 8$\times$H100 GPUs. We train for 1 epoch on AeSlides-7k-train. For each sample, we generate 8 rollouts to perform GRPO-style advantage estimation. The rollout temperature is set to 1. The maximum response length is 8192 tokens. Additional training details and hyperparameters are provided in Appendix~\ref{appx:exp-setup}, and the full training dynamics are provided in Appendix~\ref{appx:training-dynamics}.

\textbf{Evaluation}. We evaluate on AeSlides-7k-eval using the following metrics:
(i) \textit{Rendering error rate}, measuring the proportion of generated slides that fail to render. 
(ii) \textit{Verifiable aesthetic metrics} introduced in Section~\ref{sec:method}. For aspect ratio, we report size compliance under tolerances of 1\% and 5\% with respect to the standard 16:9 ratio. For the other three categories, we report the raw scores, where higher values indicate more severe issues. 
(iii) \textit{VLM-score}, computed using an LLM-as-a-judge protocol. Specifically, we employ GPT-5.2 and GPT-5-mini to assign an overall quality score to each rendered slide on a 0--5 scale; the evaluation prompt is provided in Appendix~\ref{appx:vlm-rewarding}. 
(iv) \textit{Human evaluation}. For a subset of strong models, we further conduct human evaluation focusing on aesthetic layout quality. For each query, annotators with prior experience in slide design jointly examine outputs and assign scores from all models (anonymized and in random order), to ensure a consistent evaluation scale. Given the inherently subjective nature of the task, the inter-annotator agreement is satisfactory: approximately 70\% of score differences are within 1 point, and the intraclass correlation coefficient $\mathrm{ICC}(3,1)$~\cite{shrout1979intraclass} is around 0.7. The final score is reported as the average across annotators. Additional statistics and analysis of human evaluation results are provided in Appendix~\ref{appx:human-eval}.

\textbf{Baselines}. To accurately assess the effectiveness of AeSlides in improving aesthetic slide generation, we consider four groups of baselines. (i) \textit{Proprietary models}: Claude-Sonnet-4.5 and GPT-5.2, which generate slides for the same set of prefix-conditioned queries. (ii) \textit{Base model}: GLM-4.7-Flash, i.e., the initialization model before RL training. (iii) \textit{Agentic reflection on the base model}: we adopt DeepPresenter~\cite{zheng2026deeppresenter}, a state-of-the-art open-source slide generation framework, integrated with the base model. DeepPresenter performs post-generation reflection to identify aesthetic issues and iteratively refine the slides. Since GLM-4.7-Flash does not support multimodal input, we pair it with a comparable multimodal model, GLM-4.6V-Flash~\cite{hong2025glm}, to enable heavy reflection capabilities of DeepPresenter (max reflection rounds $=10$). (iv) \textit{RL with model-based reward}: to examine the effect of verifiable rewards, we train additional models using the same RL pipeline but with model-based reward signals. We employ two reward models (GPT-5-mini and GPT-5-nano; GPT-5.2 or larger models are excluded due to prohibitive training cost, see Table~\ref{tab:meta-eval}) and consider two reward formulations: scoring static HTML directly, and scoring visual slides.


\subsection{Main Results}

We report the main experimental results in Table~\ref{tab:main-results}. Additional results, such as human evaluation win rates, Bradley-Terry scores, statistical significance tests, etc., are provided in Appendix~\ref{appx:human-eval} and~\ref{appx:additional-results}. We analyze the results from the following perspectives:

\textbf{Effectiveness of RL with verifiable aesthetic rewards}.
Compared with the base model, RL with verifiable rewards yields consistent and substantial improvements across all four verifiable metrics. These gains also translate into higher VLM-based scores and human evaluation scores, matching and even slightly edging out proprietary Claude-Sonnet-4.5 (+0.12 pts, 52\% win rate). This suggests that (i) leveraging verifiable metrics as reward signals for reinforcement learning is both effective and generalizable, and (ii) these metrics are well aligned with overall quality assessments.

\textbf{Comparison with agentic approaches}.
Compared to the base model, DeepPresenter shows no performance gains and instead consistently degrades across all metrics and human evaluation. A closer inspection reveals that approximately 70\% of samples terminate after only a single round of reflection, as the VLM prematurely concludes that the output satisfies the aesthetic criteria, despite evident layout issues. In contrast, about 0.4\% of samples exhaust all 10 reflection rounds without reaching a satisfactory solution. We attribute this to two factors: (i) the reflection mechanism in DeepPresenter primarily targets content-oriented quality and global coherence; while it can resolve severe failures (e.g., broken layouts), finer-grained aesthetic properties are still constrained by the capability of the backbone;
(ii) current VLMs fail to provide sufficiently precise and actionable visual feedback for iterative refinement of subtle aesthetic details (see ``Limitations of VLM-based scoring'' below). In addition, the agentic pipeline increases prompt tokens by 155\% and completion tokens by 46\%, introducing substantial computational overhead. Overall, we find that improving slide aesthetics currently relies more effectively on explicit aesthetic supervision during training, rather than additional agentic engineering.

\textbf{Comparison with model-based rewards}.
With sufficient capacity, the reward model can provide modest improvements in overall aesthetic quality, but fails to deliver consistent gains across all dimensions. Static rewards are effective for dimensions with well-defined rules (e.g., aspect ratio) but less sensitive to visually grounded properties. In contrast, visual rewards better capture perceptible defects such as excessive whitespace and element collision, yet remain ineffective for aspects that are difficult for current models to perceive (e.g., aspect ratio). Moreover, model-based rewards exhibit signs of reward hacking. As shown in Table~\ref{tab:main-results}, the \textit{Visual GPT-5-mini Reward} variant performs worse than \textit{Verifiable Rewards} across all metrics, yet receives higher scores from GPT-5-mini (i.e., the reward model), while GPT-5.2 and human evaluation indicate the opposite. This discrepancy highlights a key risk: while reward hacking also occurs with verifiable metrics, it is typically diagnosable and correctable via iterative refinement (as discussed in Section~\ref{sec:metrics}); in contrast, model-based rewards are more opaque, often introducing systematic bias or optimization bottlenecks.

\textbf{Limitations of VLM-based scoring}.
Table~\ref{tab:main-results} further shows that current VLMs are unreliable slide evaluators. GPT-5.2 assigns overly high scores to its own generations despite clear defects (e.g., distorted aspect ratios and element collisions), consistent with preference leakage/bias observed in prior LLM-as-a-judge studies~\cite{wataoka2024self, panickssery2024llm, lipreference}. We further compare GPT-5.2 with human annotators on the same samples and observe substantial discrepancies: the agreement is low (Quadratic Weighted Kappa $\approx 0.19$), and the correlation is weak (Spearman $\approx 0.22$). Notably, GPT-5.2 exhibits a strong regression-to-the-mean tendency, assigning conservative scores to a wide range of slides. This leads to two issues: (i) as an evaluation metric, VLM-scores lack discriminative power, causing distinct samples to receive similar or even inverted rankings; (ii) as a reward signal for RL, they exhibit insufficient intra-group variance. Empirically, in our experiments, with $G=8$, approximately 30\% of rollout groups in the \textit{Visual GPT-5-mini Reward} setting exhibit zero reward variance and are thus discarded, whereas this ratio is only 0.14\% for \textit{Verifiable Rewards w/ GDPO}. Combined with the meta-evaluation in Table~\ref{tab:meta-eval}, the untraceable reward hacking risk discussed above, and known perceptual blind spots in related studies~\cite{fu2024blink, tong2024eyes}, these results suggest that current VLMs are inadequate as evaluators or reward sources for slide generation.

\begin{table}[ht]
\centering
\small
\scalebox{0.9}{
\begin{threeparttable}
\begin{tabular}{lccccccc}
\toprule
Ablation    & A.R. & E.W. & E.C. & V.I. & R. & Ent. & VLM \\\midrule
Base Model    & 36\%     & 0.059     & 0.126     & 1.545     &  2.94  & 0.095    & 4.65    \\
Full Method    & 85\%     & 0.033     & 0.072     & 1.111     & 3.35   & 0.099     & 4.77    \\ \midrule
w/o A.R. Reward    & \textcolor{red}{39\%}     & 0.042     & 0.128     & 0.948    & -   & 0.096     &  4.58   \\
w/o E.W. Reward    & 84\%     & \textcolor{red}{0.093}     & 0.081     & 1.275    & -   & 0.100     & 4.62    \\
w/o E.C. Reward    & 77\%     & 0.044     & \textcolor{red}{0.200}     & 1.023     & -   & 0.104     & 4.60    \\
w/o V.I. Reward    & 86\%     & 0.038     & 0.090     & \textcolor{red}{2.289}     & -   & 0.096     & 4.67    \\ \midrule
w/o KL Div. & \textcolor{red}{95\%}   & \textcolor{red}{<0.01}     & \textcolor{red}{0.041}     & \textcolor{red}{0.235}     & \textcolor{red}{3.89}   & \textcolor{red}{0.072} & 4.72    \\ \midrule
w/o Shaping & 84\%     & 0.039     & 0.136     & 1.277     & \textcolor{red}{3.07}   & 0.095     & 4.57    \\
w/o GDPO    & 75\%     & 0.032     & 0.087     & 1.071     & \textcolor{red}{3.25}   & 0.098     & 4.73    \\
w/o GDPO + Shaping    & 77\%     & 0.068     & 0.144     & 0.963     & \textcolor{red}{3.21}   & 0.105     & 4.56    \\ \bottomrule
\end{tabular}
\end{threeparttable}
}
\caption{\label{tab:ablation} Ablation results. A.R.: Aspect ratio compliance under tolerance of 5\%; E.W.: excessive whitespace; E.C.: element collision; V.I.: visual imbalance. R.: total reward; for \textit{w/o shaping}, the shaped reward is reported. Ent.: policy entropy.}
\end{table}

\subsection{Ablation Studies}

To further validate the contribution of components in AeSlides, we conduct additional ablation studies, with results reported in Table~\ref{tab:ablation}.

\textbf{Ablation on reward components}.
We perform individual ablations on all four reward components. As shown in the table, removing any single component leads to a significant degradation in its corresponding metric. We also observe that certain metrics exhibit synergistic effects during optimization; for instance, optimizing the whitespace reward can indirectly improve visual imbalance.

\textbf{Ablation on KL divergence}.
We remove the KL divergence constraint and observe that the model attains seemingly high rewards. However, a sharp drop in policy entropy indicates that this comes at the cost of a rapid collapse of the policy space. As training progresses, the various design patterns learned during the SFT stage are gradually abandoned, with only the most conservative ones retained, causing the model to degenerate into a template-based generation process. This behavior is clearly undesirable, indicating that maintaining KL divergence is necessary for slide generation. Relevant failure cases and analysis are provided in Appendix~\ref{appx:kl-ablation}.

\textbf{Ablation on reward shaping and normalization}.
Both reward shaping and reward-decoupled normalization (GDPO) facilitate continuous optimization across all reward components, preventing the collapse of individual reward signals. Removing either component causes the model to over-optimize certain rewards at the expense of others, rather than achieving coordinated improvement. Human evaluation results in Table~\ref{tab:main-results} also cross-validate that GDPO contributes positively to overall quality.

\begin{figure*}[ht]
    \centering
    \includegraphics[width=0.9\linewidth]{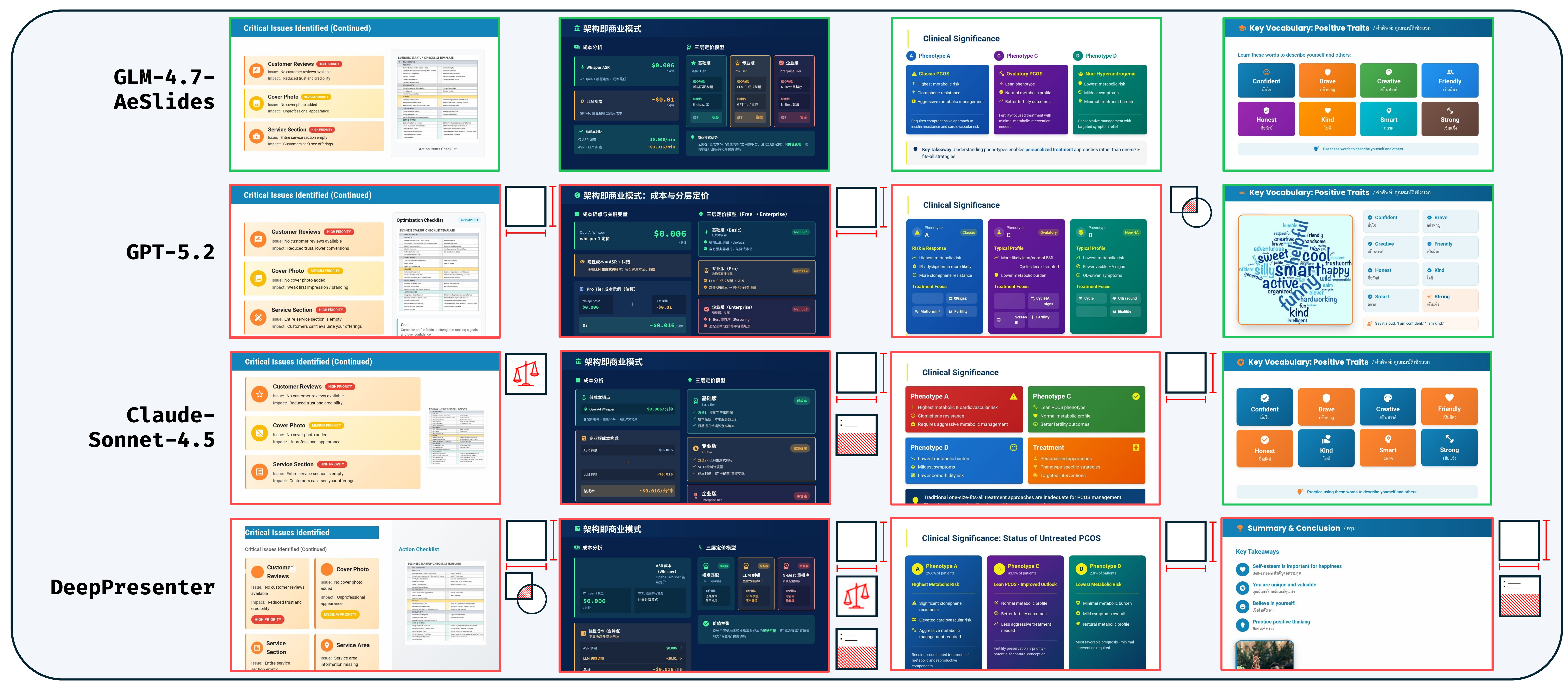}
    \caption{\label{fig:case-study}Case studies of slides generated by: AeSlides, GPT-5.2, Claude-Sonnet-4.5, and DeepPresenter. Corresponding aesthetic issues are presented on the right:
    \includegraphics[height=0.90em]{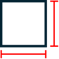} denotes distorted aspect ratio, \includegraphics[height=0.90em]{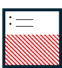} denotes excessive whitespace, \includegraphics[height=0.90em]{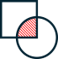} denotes element collision, and \includegraphics[height=0.90em]{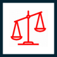} denotes visual imbalance. Slides with distorted aspect ratios are truncated to standard 16:9 for better visualization.}
\end{figure*}

\subsection{Case Studies}

We present in Figure~\ref{fig:case-study} four groups of queries and the corresponding slides generated by four different models. A consistent observation is that, due to the lack of explicit size supervision, distorted aspect ratio emerges as the most prominent issue. Human annotators also report that this factor has the largest impact on overall quality assessment.
Group 1 considers scenarios where two distinct types of content must be presented within a single slide. AeSlides accurately controls the proportion between the cards and the image, achieving both compliance with aspect ratio constraints and a well-balanced visual center. In contrast, other models fail to satisfy these two requirements simultaneously.
Group 2 focuses on cases requiring the presentation of multiple dimensions of information. AeSlides adopts a flexible layout strategy, allocating space according to content quantity, resulting in well-aligned cards and efficient space utilization. Other models either rely on simple two-column layouts that exceed height constraints or attempt more complex layouts but fail to distribute space properly, leading to excessive whitespace and visual imbalance.
Group 3 demonstrates AeSlides' capability in handling hierarchical text with appropriate content structuring and trimming, along with effective icon usage. In comparison, GPT-5.2 introduces incorrect icons that lead to rendering misalignment and subsequent element collisions. The remaining models fail to properly truncate and organize content, resulting in distorted aspect ratio.
Group 4 highlights a notable failure case of DeepPresenter. Despite obvious issues such as severe height overflow and excessive whitespace, the VLM-based evaluator fails to identify these problems and incorrectly considers the layout acceptable. This further suggests that VLMs exhibit clear limitations in perceiving certain aesthetic dimensions of slide design.
Overall, the results show that AeSlides effectively internalizes aesthetic criteria into model parameters, thereby mitigating the modality gap between text-based generation and visually grounded evaluation.

\section{Limitations and Future Work}

In this paper, we focus on four categories of aesthetic layout issues in slide generation. However, AeSlides-Reward-Bench also covers additional quality factors beyond layout, including insufficient content density, mismatch with user intent, suboptimal color schemes, and unsuitable image selection, etc. These aspects are not explicitly modeled in our current framework and are left for future work.
Moreover, the element collision and visual imbalance metrics rely on iterative refinement of heuristic rules. This dependence not only incurs substantial engineering overhead but also limits scalability. Future work should further investigate non-heuristic but verifiable metrics, such as the vision-based whitespace detection introduced in this paper.
Finally, while the proposed aesthetic reward aligns with majority preferences, it implicitly assumes a shared notion of visual quality. In practice, aesthetic standards vary across cultures and user groups, and a single reward formulation may not capture such diversity. Developing personalized or adaptive aesthetic fine-tuning remains another important direction for future work.

\section{Conclusion}
In this paper, we conduct an investigation of the aesthetic deficiencies in LLM-based slide generation, arising from the modality gap between text-based generation and visually grounded evaluation.
We first introduce \textbf{AeSlides-Reward-Bench}, a dataset with annotations over 21 categories of slide issues, and show that the dominant failure modes stem from insufficient aesthetic layout capabilities of current models.
To address this, we develop a suite of verifiable metrics that target four major categories of aesthetic layout issues. Meta-evaluations demonstrate that these metrics significantly outperform VLM-based approaches in terms of accuracy, efficiency, and cost.
Building upon these metrics, we propose \textbf{AeSlides}, a reinforcement learning framework that fine-tunes LLMs via verifiable aesthetic rewards, enabling the generation of slides with visually coherent layouts.
Extensive experiments show that, with a limited amount of data, AeSlides substantially improves the aesthetic quality of generated slides. It consistently outperforms iterative reflection methods and model-based reward optimization across all metrics and human evaluations, and even edges out proprietary systems such as Claude-Sonnet-4.5.
Further analysis reveals the unreliability of current VLMs in evaluating and supervising slide generation, underscoring the importance of verifiable reward design for improving aesthetic slide generation.
Code, datasets, and model are publicly released to facilitate future research.

\bibliographystyle{ACM-Reference-Format}
\bibliography{ref}

\appendix
\section{Metadata and Statistics of Datasets and Model Checkpoints}
\label{appx:data-meta}

\subsection{AeSlides-Reward-Bench}

\begin{table}[htbp]
\centering
\small
\begin{threeparttable}
  \scalebox{0.92}{
\begin{tabular}{ll} 
  \toprule
Metadata Field                         & Value     \\\midrule
License                                & CC BY 4.0 \\
Annotation Date                        & Dec. 2025 \\
Annotator Count                        & 6         \\
Issue Category Count                   & 4         \\
Issue Subcategory Count                & 21        \\
Total Annotation Effort (person-hours) & $\sim$300          \\ \midrule
Record Count                           & 2068      \\
Paired Task Count                      & 1029      \\
Singleton Task Count                   & 10        \\
Query Length (min, median, mean, max)                     & 16, 608, 3017.8, 42412        \\
Preceding Slides Count (min, median, mean, max)          & 0, 4, 4.93, 24      \\
\bottomrule
\end{tabular}
  }
\end{threeparttable}
\caption{\label{tab:reward-bench-meta} Metadata and statistics of AeSlides-Reward-Bench.}
\end{table}

\begin{table*}[t]
\centering
\small
\begin{threeparttable}
\scalebox{0.9}{
\begin{tabular}{c|l|l|c|c}
  \toprule
Category                 & Subcategory                        & Description                                                                                       & \# Labels & Defect \% \\ \midrule
\multirow{7}{*}{Layout}  & {[}L-1{]} Distorted Aspect Ratio   & Compared to the standard 16:9 aspect ratio, the aspect ratio is excessively wide or narrow.       & 3           & 56.0      \\
                         & {[}L-2{]} Missing Centering        & Elements that should be centered (e.g., the main title) are not centered.                         & 4           & 12.5      \\
                         & {[}L-3{]} Misalignment             & Elements that should be aligned (e.g., peer-level cards) are not properly aligned.                & 4           & 6.0       \\
                         & {[}L-4{]} Excessive Whitespace     & Large areas of empty space exist on the page, resulting in a visually hollow appearance.          & 4           & 46.5      \\
                         & {[}L-5{]} Content Overcrowding     & Content is overly crowded or consists entirely of dense text, making it difficult to read.        & 3           & 0.9       \\
                         & {[}L-6{]} Visual Imbalance         & Poor layout structure, with unbalanced spacing and disharmonious density distribution.            & 3           & 25.0      \\
                         & {[}L-7{]} Element Collision        & Elements overlap, overflow their parent container, or exceed slide boundaries.                    & 2           & 13.9      \\ \midrule
\multirow{5}{*}{Vision}  & {[}V-1{]} Low Contrast             & Color contrast is too weak, making text difficult to read.                                        & 3           & 1.4       \\
                         & {[}V-2{]} Jarring Contrast         & Color contrast is too strong, appearing harsh or visually unappealing.                            & 2           & 5.4       \\
                         & {[}V-3{]} Cluttered Color Scheme   & Too many colors are used, resulting in a cluttered and chaotic presentation.                      & 2           & 3.2       \\
                         & {[}V-4{]} Inconsistent Style       & The color scheme of the current page differs significantly from previous pages.                   & 2           & 1.3       \\
                         & {[}V-5{]} Lack of Style            & Insufficient styling, such as missing hierarchy or lack of typographic structure in large blocks. & 2           & 0.1       \\ \midrule
\multirow{7}{*}{Element} & {[}E-1{]} Irrational Image Size    & Image sizes are inappropriate, leading to poor arrangement or an overall unattractive layout.     & 3           & 16.5      \\
                         & {[}E-2{]} Irrational Image Content & Image content is inappropriate, either inconsistent with the theme or visually unappealing.       & 4           & 80.8      \\
                         & {[}E-3{]} Poor Image Cropping      & Image cropping is improper.                                                                       & 3           & 4.5       \\
                         & {[}E-4{]} Poor Vector Graphics     & Vector graphics or geometric elements are poorly constructed, resulting in poor visual quality.   & 3           & 8.3       \\
                         & {[}E-5{]} Poor Charts/Tables       & Tables/charts exhibit misalignment, structural errors, or excessive length.                       & 3           & 4.0       \\
                         & {[}E-6{]} Inappropriate Font Size  & Such as children significantly larger than parents, or inconsistent sizes among peers.            & 5           & 4.3       \\
                         & {[}E-7{]} Missing Icons            & Icons are missing, misaligned, or occupy unnecessary space.                                       & 2           & 4.5       \\ \midrule
\multirow{2}{*}{Content} & {[}C-1{]} Irrelevant Content       & The generated page content does not match user requirements.                                      & 2           & 0.1       \\
                         & {[}C-2{]} Low Information Density  & The page contains too little content, resulting in a sparse layout.                               & 3           & 1.4       \\ \midrule
\multicolumn{2}{c|}{Overall Preference (Pairwise)}             & Overall preference between two different slide pages generated from the same query.               & 3           & -        \\ \bottomrule
\end{tabular}
}
\end{threeparttable}
\caption{\label{tab:reward-bench-criteria}  The annotation criteria of AeSlides-Reward-Bench span four dimensions (\textit{Layout}, \textit{Vision}, \textit{Element}, and \textit{Content}) with 21 subcategories. Detailed labels are listed in Table~\ref{tab:reward-bench-labels}.}
\end{table*}

\begin{table*}[t]
\centering
\small
\begin{threeparttable}
\scalebox{0.92}{
\begin{tabular}{ll}
  \toprule
Subcategory                        & Labels                                                                                                                      \\ \midrule
{[}L-1{]} Distorted Aspect Ratio   & \textcolor{red}{Excessive aspect ratio}; \textcolor{red}{Insufficient aspect ratio}; \textcolor{ForestGreen}{No issue}.          \\
{[}L-2{]} Missing Centering        & \textcolor{red}{Severe issue}; \textcolor{red}{Minor issue}; \textcolor{ForestGreen}{Proper centering}; Not applicable.           \\
{[}L-3{]} Misalignment             & \textcolor{red}{Severe issue}; \textcolor{red}{Minor issue}; \textcolor{ForestGreen}{Proper alignment}; Not applicable.           \\
{[}L-4{]} Excessive Whitespace     & \textcolor{red}{Severe issue}; \textcolor{red}{Minor issue}; \textcolor{ForestGreen}{No issue}; Not applicable.                   \\
{[}L-5{]} Content Overcrowding     & \textcolor{red}{Overcrowding}; \textcolor{ForestGreen}{No issue}; Not applicable.                                                  \\
{[}L-6{]} Visual Imbalance         & \textcolor{red}{Imbalance}; \textcolor{ForestGreen}{No issue}; Not applicable.                                                     \\
{[}L-7{]} Element Collision        & \textcolor{red}{Element collision}; \textcolor{ForestGreen}{No issue}.                                                             \\ \midrule
{[}V-1{]} Low Contrast             & \textcolor{red}{Globally low contrast}; \textcolor{red}{Locally low contrast}; \textcolor{ForestGreen}{No issue}.                 \\
{[}V-2{]} Jarring Contrast         & \textcolor{red}{Excessive contrast}; \textcolor{ForestGreen}{No issue}.                                                            \\
{[}V-3{]} Cluttered Color Scheme   & \textcolor{red}{Cluttered color scheme}; \textcolor{ForestGreen}{No issue}.                                                        \\
{[}V-4{]} Inconsistent Style       & \textcolor{red}{Style inconsistency}; \textcolor{ForestGreen}{No issue}.                                                           \\
{[}V-5{]} Lack of Style            & \textcolor{red}{Lack of stylistic coherence}; \textcolor{ForestGreen}{No issue}.                                                   \\ \midrule
{[}E-1{]} Irrational Image Size    & \textcolor{red}{Improper image sizing}; \textcolor{ForestGreen}{Proper image sizing}; Not applicable.                             \\
{[}E-2{]} Irrational Image Content & \textcolor{red}{Inappropriate background image}; \textcolor{red}{Inappropriate illustrative image}; \textcolor{ForestGreen}{Proper image usage}; Not applicable. \\
{[}E-3{]} Poor Image Cropping      & \textcolor{red}{Poor image cropping}; \textcolor{ForestGreen}{Proper image cropping}; Not applicable.                             \\
{[}E-4{]} Poor Vector Graphics     & \textcolor{red}{Poor vector graphic quality}; \textcolor{ForestGreen}{Proper vector graphic usage}; Not applicable.               \\
{[}E-5{]} Poor Charts/Tables       & \textcolor{red}{Poor chart/table quality}; \textcolor{ForestGreen}{Proper chart/table usage}; Not applicable.                     \\
{[}E-6{]} Inappropriate Font Size  & \textcolor{red}{Child font larger than parent}; \textcolor{red}{Inconsistent sibling font sizes}; \textcolor{red}{Overall font too small}; \textcolor{red}{Title font excessively large}; \textcolor{ForestGreen}{Proper font size}. \\
{[}E-7{]} Missing Icons            & \textcolor{red}{Missing icons}; \textcolor{ForestGreen}{No issue}.                                                                 \\ \midrule
{[}C-1{]} Irrelevant Content       & \textcolor{red}{Content mismatch}; \textcolor{ForestGreen}{No issue}.                                                              \\
{[}C-2{]} Low Information Density  & \textcolor{red}{Low information density}; \textcolor{ForestGreen}{No issue}; Not applicable.                                      \\ \bottomrule
\end{tabular}
}
\end{threeparttable}
\caption{\label{tab:reward-bench-labels} The label taxonomy across dimensions in AeSlides-Reward-Bench: \textcolor{ForestGreen}{Green} indicates OK, \textcolor{red}{Red} indicates Defect, and Black denotes Not Applicable. }
\end{table*}

\subsubsection{Annotation Protocol}

AeSlides-Reward-Bench is introduced to provide meta-evaluation of diverse slide issues, beyond the four layout issues studied in this work. It covers four dimensions (\textit{Layout}, \textit{Vision}, \textit{Element}, and \textit{Content}) with 21 subcategories. The corresponding annotation criteria are listed in Table~\ref{tab:reward-bench-criteria} (annotators are provided with Chinese translations).
We define three primary labels: OK (no issue), Defect (issue present), and Not Applicable (e.g., no relevant elements or dimensions excluded for special pages). For certain cases, finer-grained labels are used (e.g., minor vs. severe issue for excessive whitespace). The full label taxonomy is shown in Table~\ref{tab:reward-bench-labels}.
During annotation, annotators are given: (i) the user request and supporting context, (ii) preceding slides, and (iii) two rollout candidates from the same query. They score each candidate across all 21 dimensions and then provide an overall preference indicating which candidate is superior in overall quality.

\subsubsection{Statistics}

Table~\ref{tab:reward-bench-criteria} also reports the defect percentage for each dimension. 
Additional metadata and statistics of the dataset are summarized in Table~\ref{tab:reward-bench-meta}.

\subsubsection{Scope and Justification of Selected Layout Issues}
Although AeSlides-Reward-Bench annotates 21 issue dimensions, this work focuses on four layout-related issues for two reasons.
First, using the annotated overall preference, we perform Bradley-Terry-style modeling at the category level. For each subcategory, labels are mapped to $(-1, 0, 1)$, averaged within each category, and the difference between paired candidates is used as the relative strength to fit the model, yielding 
ROC-AUC$ \approx 0.85$. Among the four categories, layout issues contribute the most to overall preference, with a weight of 4.43, compared to 2.49, 1.60, and 1.61 for Vision, Element, and Content, respectively. This indicates that layout issues are currently the primary bottleneck.
Second, we conduct a finer-grained Bradley-Terry analysis over all 21 subcategories. Within the layout category, L-4, L-6, and L-7 (corresponding to excessive whitespace, visual imbalance, and element collision) exhibit the highest weights. In addition, L-1 (distorted aspect ratio) has the highest defect percentage. We therefore select these four dimensions as the main focus of this study.
Based on these findings, we target slide quality improvement through these four layout aesthetic dimensions, while leaving the remaining dimensions as future work.

\subsection{AeSlides-7k}

\begin{table}[htbp]
\centering
\small
\begin{threeparttable}
\begin{tabular}{ll}
  \toprule
Metadata Field                                & Value             \\ \midrule
Sample Count (total)                          & 8471              \\
Filtered Special Sample Count                 & 1735              \\
Sample Count (train, eval)                    & 5704, 1032        \\
Prompt Tokens (min, median, mean, max)        & 8k, 22k, 21k, 30k \\
Input Messages Count (min, median, mean, max) & 6, 24, 27, 84     \\
Language Count                                & \textgreater{}5  \\ \bottomrule
\end{tabular}
\end{threeparttable}
\caption{\label{tab:7k-meta} Metadata and statistics of AeSlides-7k.}
\end{table}

\begin{figure}[htbp]
    \centering
    \includegraphics[width=0.95\linewidth]{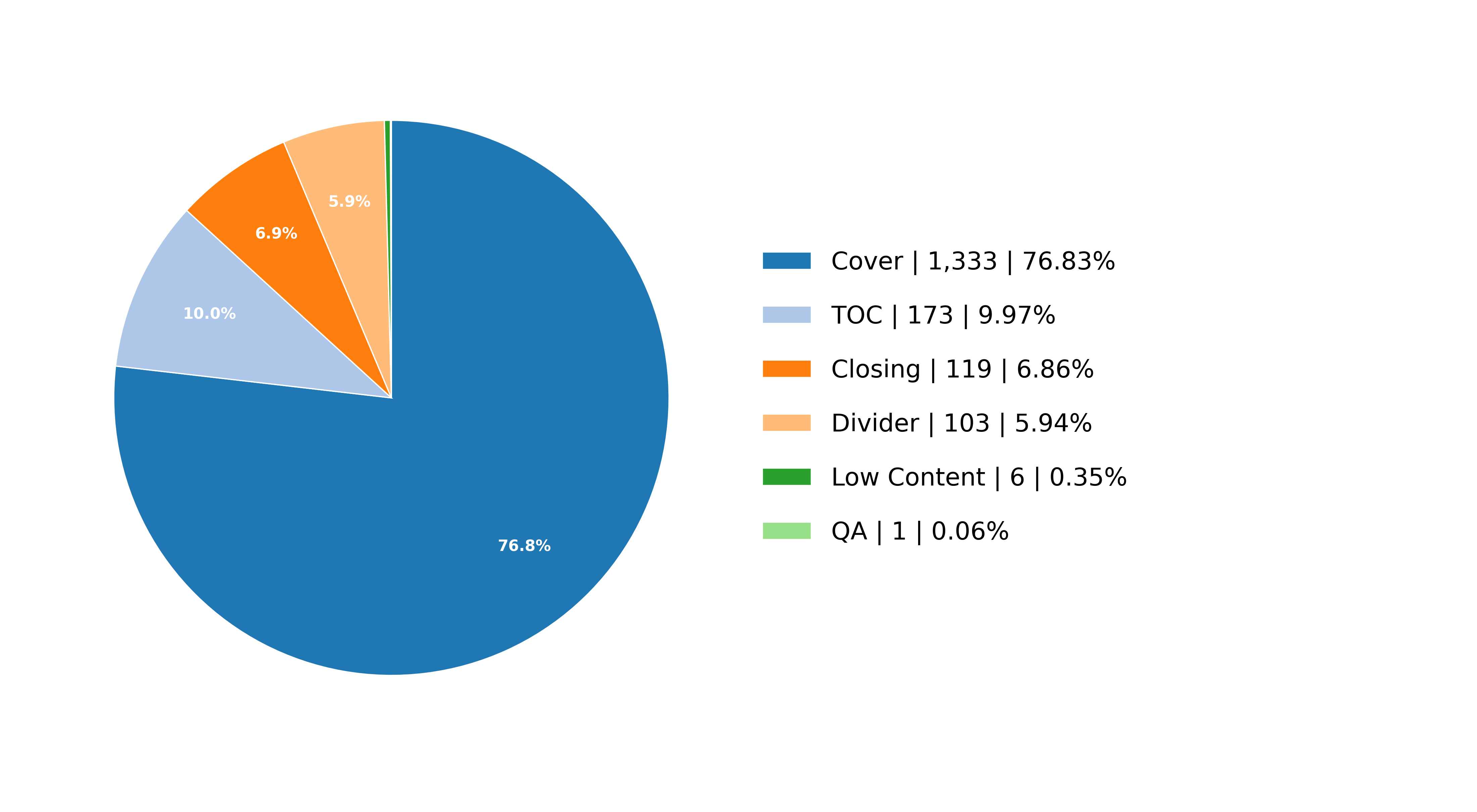}
    \caption{Distribution of filtered structurally simple pages in AeSlides-7k.}
    \label{fig:pie-filter-page}
\end{figure}

\begin{figure}[htbp]
    \centering
    \includegraphics[width=0.95\linewidth]{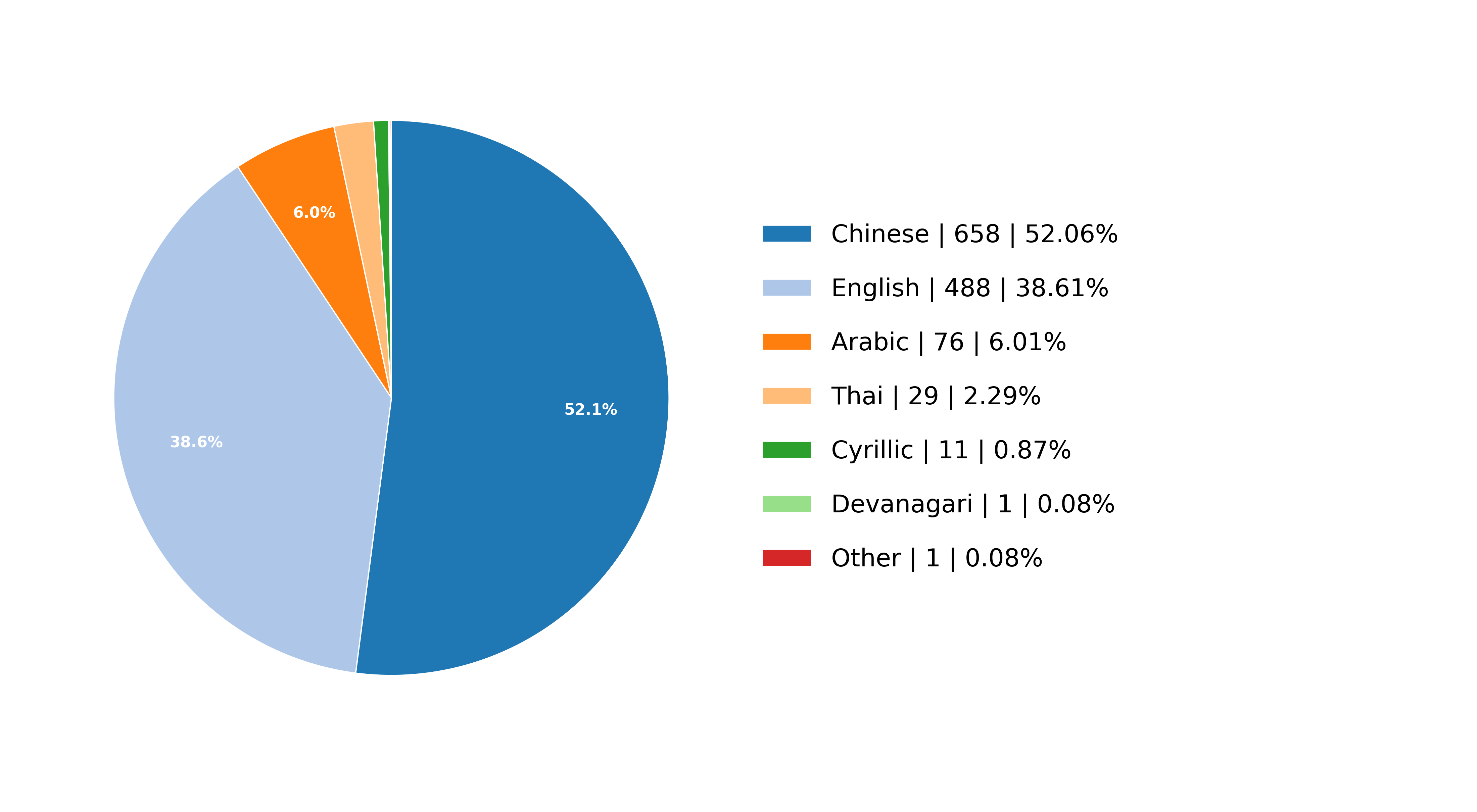}
    \caption{Distribution of languages in AeSlides-7k.}
    \label{fig:pie-language}
\end{figure}

\begin{figure}[htbp]
    \centering
    \includegraphics[width=0.95\linewidth]{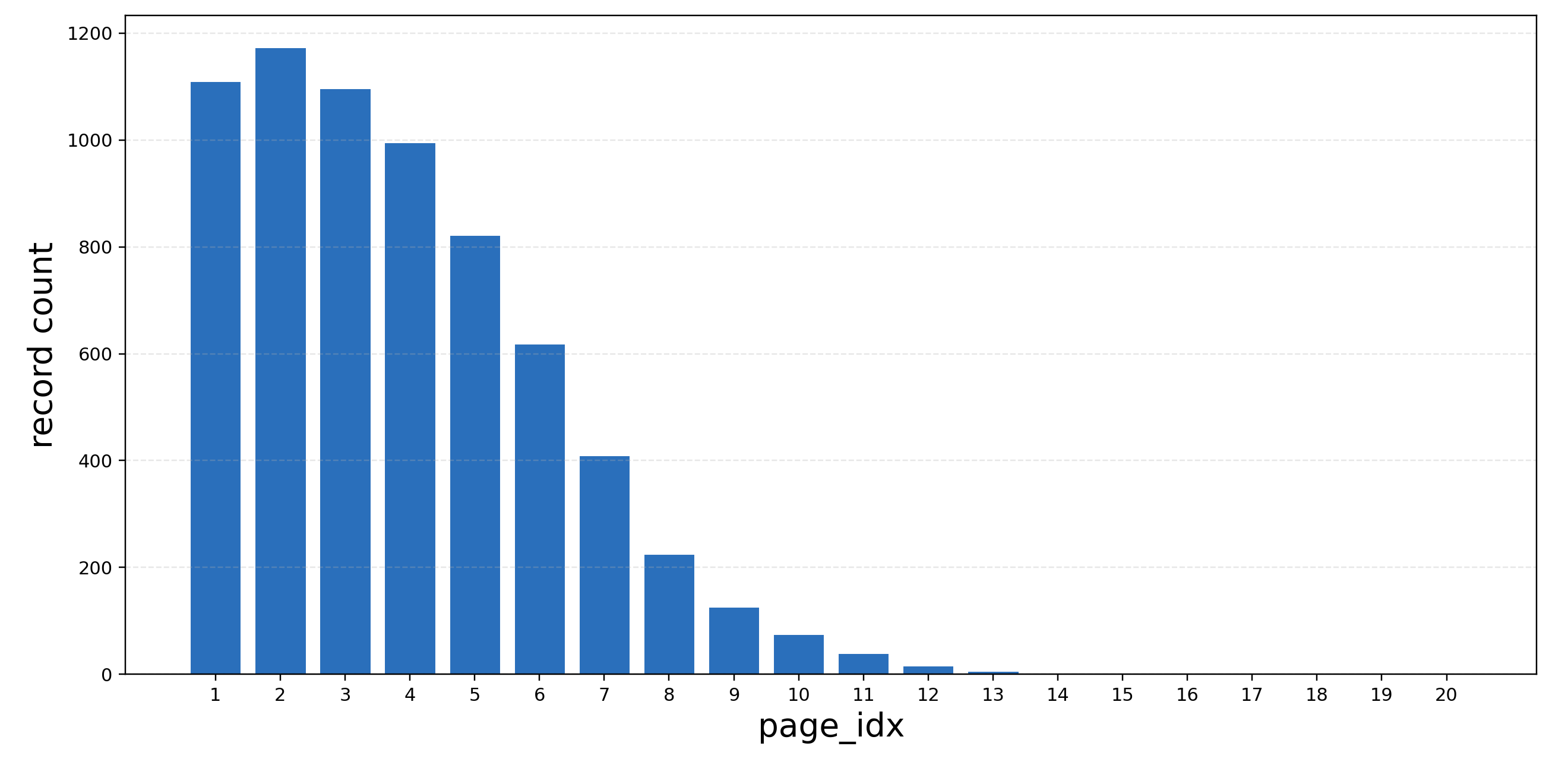}
    \caption{Distribution of page indices in AeSlides-7k.}
    \label{fig:page-idx}
\end{figure}

\begin{figure}[htbp]
    \centering
    \includegraphics[width=0.95\linewidth]{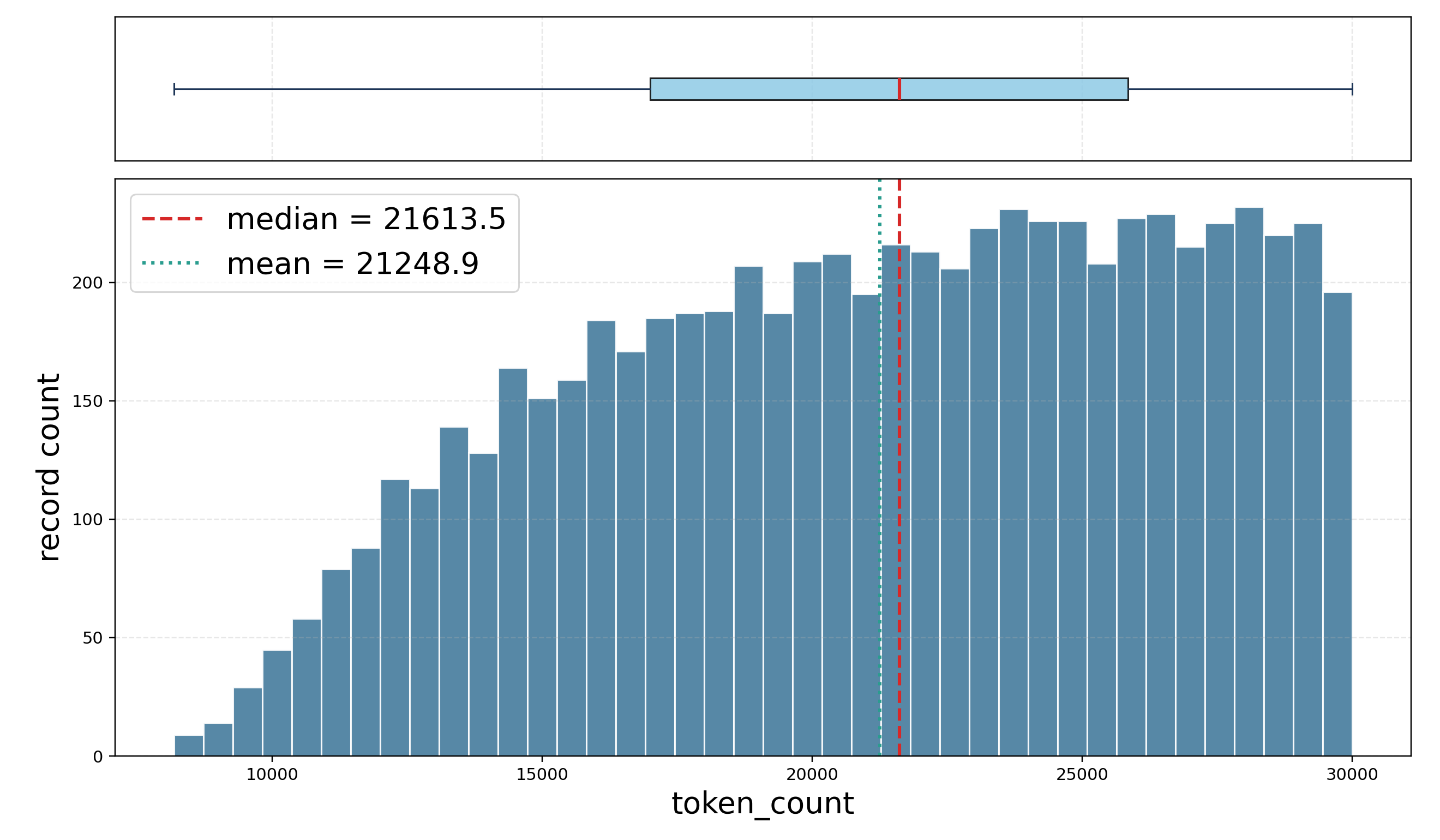}
    \caption{Distribution of prefix token counts in AeSlides-7k.}
    \label{fig:token-count}
\end{figure}

\begin{figure}[htbp]
    \centering
    \includegraphics[width=0.95\linewidth]{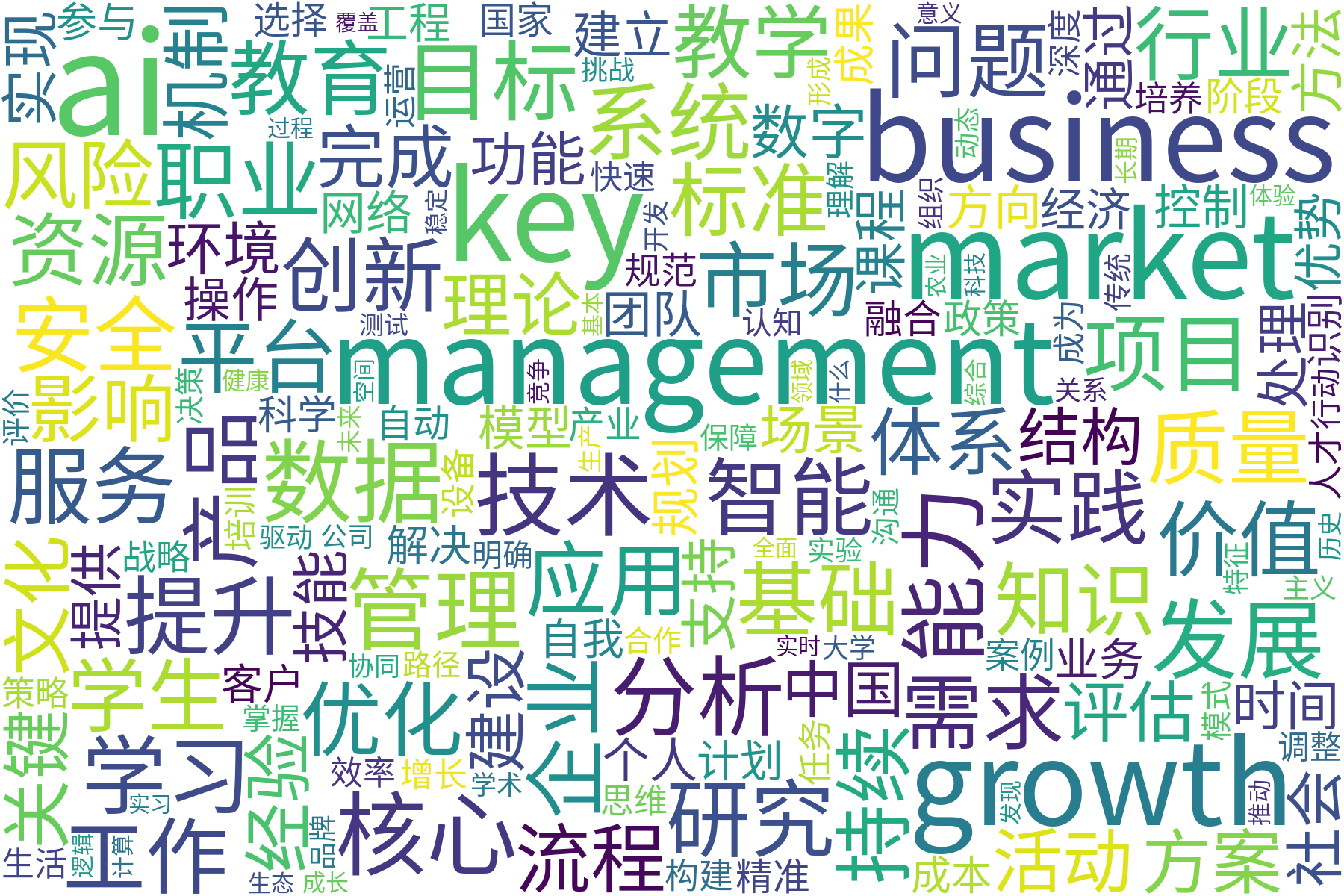}
    \caption{Word cloud of visible text in AeSlides-7k.}
    \label{fig:wordcloud}
\end{figure}

\subsubsection{Statistics}
To enable reinforcement learning to internalize aesthetic criteria, we further construct the AeSlides-7k dataset, which serves as the prefix prompts during the rollout stage of reinforcement learning. Table~\ref{tab:7k-meta} summarizes its metadata and statistics.
As described in the main paper, we filter out most structurally simple pages (e.g., cover pages, tables of contents, dividers, and ending pages). The proportions of filtered page types are visualized in Figure~\ref{fig:pie-filter-page}.
We also analyze the multilingual distribution of AeSlides-7k, which covers at least 5 languages, including English, Chinese, Arabic, Thai, Cyrillic. The language distribution is shown in Figure~\ref{fig:pie-language}.
In addition, we report the distributions of page indices within each query, prompt token counts, and visible text word clouds, which are visualized in Figure~\ref{fig:page-idx}, Figure~\ref{fig:token-count}, and Figure~\ref{fig:wordcloud}, respectively.

\subsubsection{Case}

We present a concrete example in Figure~\ref{fig:case-7k} to illustrate the formulation of prefix-conditioned slide generation used in the main paper. Specifically, the prefix prompt already contains all necessary supporting context, including external information retrieval, image search, user intent clarification, high-level planning, etc. During the reinforcement learning stage, the model generates the next slide conditioned on this prefix (highlighted in red), enabling clearer disentanglement and attribution of its aesthetic capabilities.

\subsection{GLM-AeSlides}

Metadata and statistics of the GLM-AeSlides checkpoints are summarized in Table~\ref{tab:ckpt-meta}.

\begin{table}[htbp]
\centering
\small
\begin{threeparttable}
\begin{tabularx}{\linewidth}{lX}
\toprule
Metadata Field & Value \\
\midrule
License & CC BY 4.0 \\
Base Model & GLM-4.7-Flash\tnote{a} \\
Total Parameter Count & 29.94B \\
Activated Parameter Count & 3.90B \\
Training Prompt Count & 5.7K \\
Training Sample Count & 46K \\
\bottomrule
\end{tabularx}
\end{threeparttable}
\begin{tablenotes}
\item \parbox[t]{\linewidth}{\textsuperscript{a}: \url{https://huggingface.co/zai-org/GLM-4.7-Flash}}
\end{tablenotes}
\caption{\label{tab:ckpt-meta} Metadata and statistics of GLM-AeSlides checkpoint.}
\end{table}

\section{Details of Cost and Efficiency Analysis}
\label{appx:cost}

To estimate the cost of VLM-based evaluation, we sampled approximately 100 prompts from the training set, used a base model (GLM-4.7-Flash) to generate one round of responses, and then submitted the outputs to the VLM for evaluation. Under our setup, each sample consumes roughly 2k prompt tokens (dominated by the rendered slide image tokens; this value is obtained from \texttt{response.usage} and thus already accounts for the multiplier mechanism of GPT-5-nano and GPT-5-mini) and 40 completion tokens. Pricing is based on the official documentation\footnote{\url{https://developers.openai.com/api/docs/models}
, accessed March 2026}. The estimated cost for 50k samples is $\sim$\$200, $\sim$\$30, and $\sim$\$6 for GPT-5.2, GPT-5-mini, and GPT-5-nano, respectively. When only an overall score is generated (as in our main experiments for VLM-based rewarding and evaluation), the number of prompt tokens slightly increases while completion tokens decrease; however, since vision input remains the dominant factor, the overall cost does not change substantially.

For efficiency, we measure the end-to-end latency of both the proposed verifiable metrics and VLM-based evaluation. VLM measurements are averaged over three separate time intervals to ensure robustness against temporal fluctuations. Both methods are integrated with the rendering infrastructure, and thus include a fixed rendering latency of approximately $\sim$3000 ms to ensure stable page rendering. Beyond this, verifiable metrics incur only lightweight metric computation and minimal JavaScript injection overhead (in total about $\sim$1000 ms). In contrast, VLM-based evaluation introduces significantly higher latency due to network transmission, remote request queuing, and server-side model inference.

\section{Prompts}
\label{appx:prompt}

\subsection{Prompt for VLM-Based Issue Detection}
\label{appx:vlm-issue-detection}

The prompt for VLM-based issue detection is shown in Figure~\ref{fig:vlm-prompt}. For readability, we apply necessary simplifications based on the original chat template (the same applies hereafter).

\subsection{Prompt for VLM Rewarding and Overall Scoring}
\label{appx:vlm-rewarding}

The prompt for VLM-based overall aesthetic evaluation is shown in Figure~\ref{fig:vlm-overall-prompt}. Its static counterpart follows a similar structure, but replaces the rendered slide screenshot with the HTML source code of the slide, with corresponding adjustments to the prompt.

\begin{figure*}[htbp]
    \centering
    \includegraphics[width=0.99\linewidth]{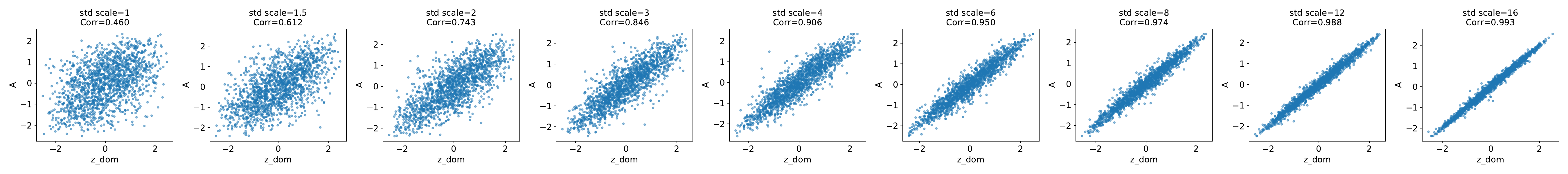}
    \caption{Scatter plot of the normalized advantage against the dominant standard deviation in Monte Carlo simulations.}
    \label{fig:theory-scatter}
\end{figure*}

\section{Theoretical Justification of Reward Signal Collapse in Multi-Reward Optimization}
\label{appx:reward-collapse}

In this section, we provide a brief theoretical justification for the reward collapse phenomenon in multi-reward reinforcement learning. In particular, GDPO~\cite{liu2026gdpo} identifies one form of reward collapse under discretized rewards: different reward combinations may map to only a few distinct advantage values after summation and normalization (e.g., $(2,0)$ and $(1,1)$ yielding the same total reward), which compresses reward composition information and weakens the optimization signal of certain components. In this work, we present an alternative perspective by analyzing reward collapse in the general setting of continuous rewards through the variance scale across reward components.

We first revisit the advantage estimation in standard GRPO. Given a group of size $G$, the model produces rollouts $y_1,\ldots,y_G$. Assume there are $K$ reward components, and denote the $k$-th reward component of sample $j$ by $r_j^{(k)}$. Standard GRPO first aggregates the reward components,
\begin{equation}
R_j = \sum_{k=1}^K r_j^{(k)},
\end{equation}
and then applies group-wise normalization,
\begin{equation}
A_j = \frac{R_j - \bar R}{\hat{\sigma}_R},
\qquad
\bar R := \frac{1}{G}\sum_{j=1}^G R_j,
\qquad
\hat{\sigma}_R^2 := \frac{1}{G}\sum_{j=1}^G (R_j-\bar R)^2.
\end{equation}

For each reward component, define the within-group sample mean and sample standard deviation as
\begin{equation}
\bar r^{(k)} := \frac{1}{G}\sum_{j=1}^G r_j^{(k)},
\qquad
\hat{\sigma}_k^2 := \frac{1}{G}\sum_{j=1}^G \left(r_j^{(k)}-\bar r^{(k)}\right)^2.
\end{equation}
Using $\frac{1}{G-1}$ instead of $\frac{1}{G}$ would only change a constant factor and does not affect the conclusion, and is thus omitted for simplicity.
Since $\bar R = \sum_{k=1}^K \bar r^{(k)}$, we have the exact finite-group decomposition
\begin{equation}
A_j
=
\frac{\sum_{k=1}^K \left(r_j^{(k)}-\bar r^{(k)}\right)}{\hat{\sigma}_R}
=
\sum_{k=1}^K
\frac{\hat{\sigma}_k}{\hat{\sigma}_R}
\cdot
\frac{r_j^{(k)}-\bar r^{(k)}}{\hat{\sigma}_k}
=
\sum_{k=1}^K w_k z_j^{(k)},
\end{equation}
where
\begin{equation}
w_k := \frac{\hat{\sigma}_k}{\hat{\sigma}_R},
\qquad
z_j^{(k)} := \frac{r_j^{(k)}-\bar r^{(k)}}{\hat{\sigma}_k}.
\end{equation}
By construction, each $z_j^{(k)}$ has zero within-group sample mean and unit within-group sample variance. Therefore, standard GRPO does not combine reward components equally after aggregation; instead, it forms a variance-weighted mixture of the per-component normalized signals, with weights determined by the within-group sample scales $w_k$.

To further interpret these weights, note that
\begin{equation}
\hat{\sigma}_R^2
=
\sum_{k=1}^K \hat{\sigma}_k^2
+
2\sum_{1\le k < \ell \le K}\hat c_{k\ell},
\end{equation}
where
\begin{equation}
\hat c_{k\ell}
:=
\frac{1}{G}\sum_{j=1}^G
\left(r_j^{(k)}-\bar r^{(k)}\right)
\left(r_j^{(\ell)}-\bar r^{(\ell)}\right)
\end{equation}
is the within-group sample covariance between reward components $k$ and $\ell$. Under the assumption that the population cross-component covariances are near zero, these sample covariance terms fluctuate around zero and are typically small relative to the variance terms for moderate $G$. In that case,
\begin{equation}
\hat{\sigma}_R^2 \approx \sum_{k=1}^K \hat{\sigma}_k^2,
\qquad
w_k \approx \frac{\hat{\sigma}_k}{\sqrt{\sum_{\ell=1}^K \hat{\sigma}_\ell^2}}.
\end{equation}

Now suppose one reward component has a much larger variance scale than the others:
\begin{equation}
\hat{\sigma}_m^2 \gg \sum_{k\neq m}\hat{\sigma}_k^2,
\end{equation}
and the sample covariance terms are not dominant. Then
\begin{equation}
w_m \approx 1,
\qquad
w_k \approx \frac{\hat{\sigma}_k}{\hat{\sigma}_m}\ll 1 \quad (k\neq m),
\end{equation}
so that
\begin{equation}
A_j
=
w_m z_j^{(m)} + \sum_{k\neq m} w_k z_j^{(k)}
\approx
z_j^{(m)}.
\end{equation}
In this regime, the normalized advantage is dominated by the high-variance component, while the contributions of the remaining components are strongly attenuated. Consequently, the policy update is driven primarily by the dominant component, and the optimization signals from the other components effectively collapse.
A practical remedy is to align reward scales through reward reweighting or reward shaping. Another solution is to apply GDPO-style reward-decoupled normalization. In this work, we employ both, and the ablation studies in the main paper support this design choice.

\begin{figure}[htbp]
    \centering
    \includegraphics[width=0.9\linewidth]{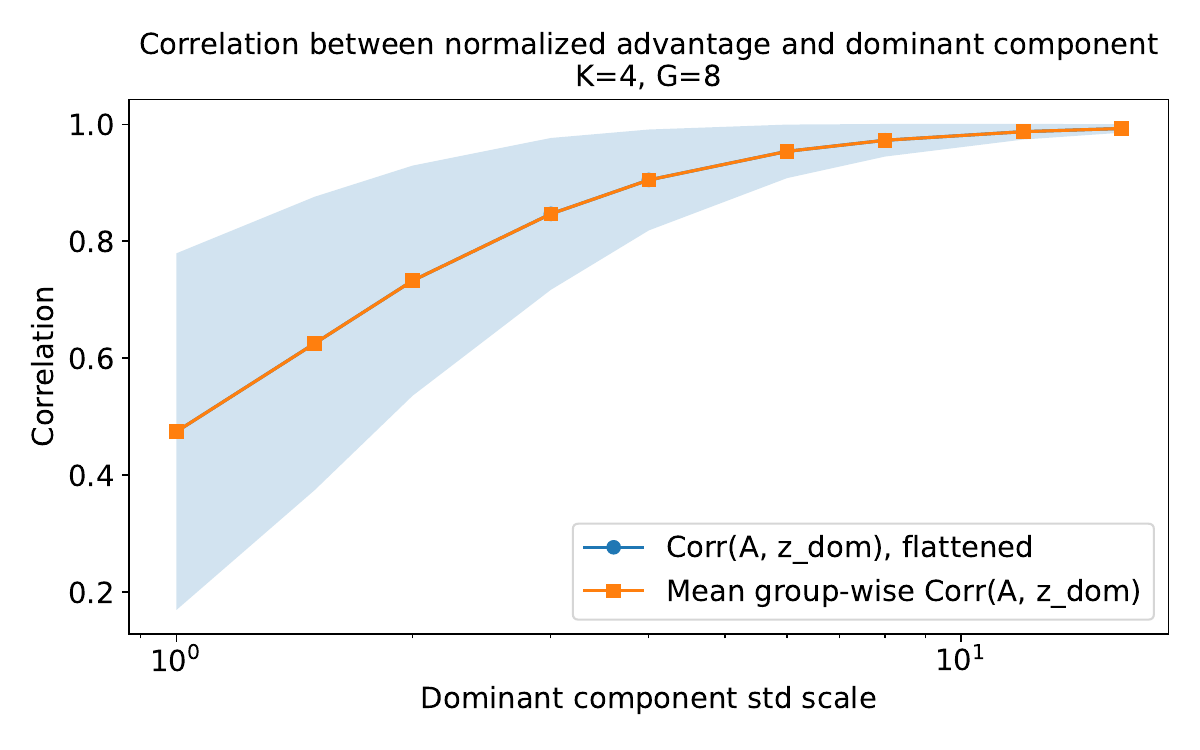}
    \caption{Correlation between the normalized advantage and the dominant standard deviation in Monte Carlo simulations.}
    \label{fig:theory-correlation}
\end{figure}

We further conduct Monte Carlo simulations to visualize this phenomenon. Under our experimental setup ($K=4, G=8$), we fix the scales of the non-dominant reward components and vary the standard deviation of one component. We then compute the correlation between the resulting normalized advantage and the normalized signal $z^{(m)}$ of the dominant component (Figure~\ref{fig:theory-correlation}), together with the corresponding scatter plots (Figure~\ref{fig:theory-scatter}). The results show that when the standard deviation of one component becomes roughly 3x larger than those of the others, it already dominates the optimization process. As this ratio further increases, the normalized advantage becomes nearly a linear function of that component.

\section{Additional Experimental Setup, Results and Analysis}

\subsection{Details of Experimental Setup}
\label{appx:exp-setup}
We provide additional details of the experimental setup in Table~\ref{tab:hyperparams}.

\begin{table}[htbp]
\centering
\small
\begin{threeparttable}
\begin{tabular}{ll}
\toprule
Hyperparameter                     & Value      \\ \midrule
\multicolumn{2}{l}{\textit{Training \& Rollout}}   \\ \midrule
Hardware                             & 8$\times$H100 GPUs \\
Num. Train Samples                       & 5704       \\
Epoch                              & 1          \\
Num. Rollouts                       & 356        \\
Rollout Batch Size                 & 16         \\
Num. Samples per Prompt (Group Size)& 8          \\
Rollout Max Response Length & 8192       \\
Rollout Temperature                & 1          \\
KL Loss Coefficient                & 0.1        \\
KL Loss Type                       & low\_var\_kl \\
$\varepsilon_\text{low}$                        & 0.2        \\
$\varepsilon_\text{high}$                       & 0.28       \\
Learning Rate                      & 1e-6       \\
Weight Decay                       & 0.1        \\
Adam Beta1                         & 0.9        \\
Adam Beta2                         & 0.98       \\ \midrule
\multicolumn{2}{l}{\textit{Evaluation}}   \\\midrule
Num. Eval Samples & 1032 \\
Eval Temperature                   & 0.6        \\
Eval Top-P                         & 0.95       \\\midrule
\multicolumn{2}{l}{\textit{Render Infrastructure \& Verifiable Metrics}}   \\\midrule
Hardware & 20$\times$24c48g CPU nodes \\
A.R. $\alpha$ & 16.0 \\
A.R. $\beta$ & 64.0 \\
A.R. margin $m$ & 0.04 \\

E.W. Gaussian Kernel & (21$\times$21) \\
E.W. Box Kernel & (201$\times$151) \\
E.W. $T_\text{clip}$ & 50.0 \\
E.W. $\tau$ & 0.05 \\
E.W. Clip Threshold (Lower \& Upper) & [0.8, 0.995] \\

E.C. Clip Threshold (Lower \& Upper) & [0.5, 0.998] \\

V.I. Clip Threshold (Lower \& Upper) & [0.1, 0.95] \\

\bottomrule
\end{tabular}
\end{threeparttable}
\caption{\label{tab:hyperparams} Additional details of the experimental setup.}
\end{table}

\begin{figure}[htbp]
    \centering
    \includegraphics[width=0.8\linewidth]{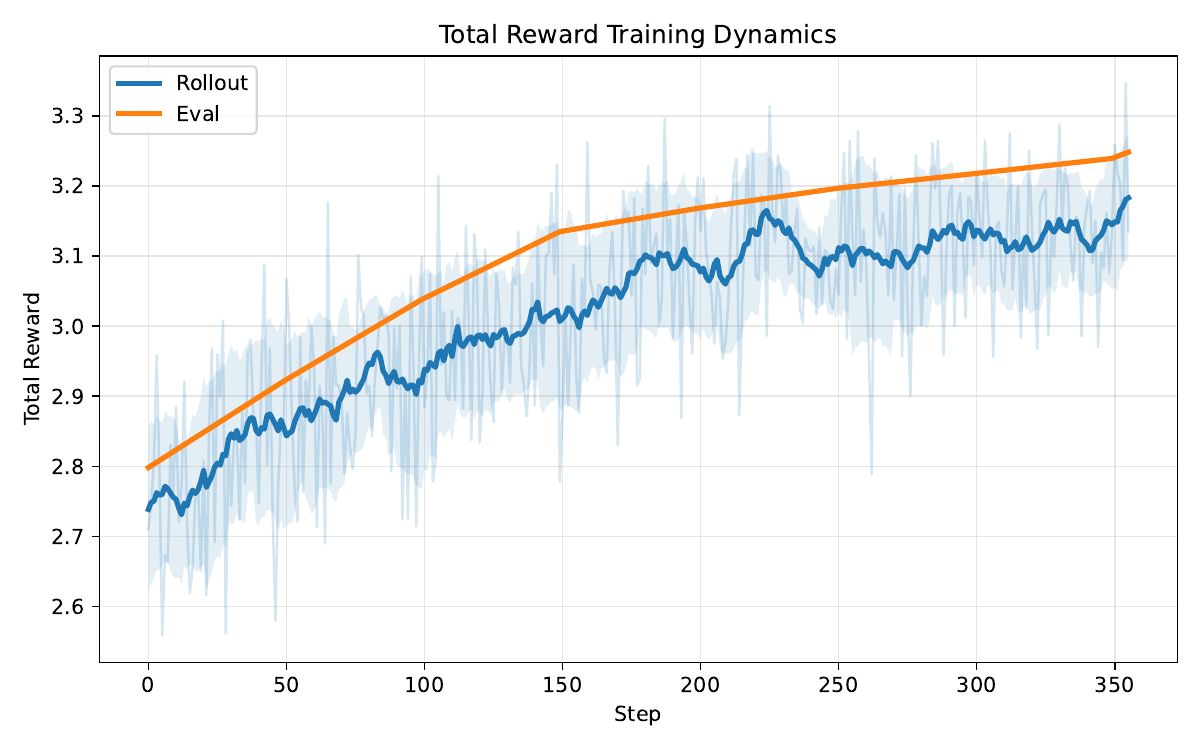}
    \caption{Reward dynamics during AeSlides training. (\textit{smoothed with a rolling window of size 30; mean ± std})}
    \label{fig:reward-dynamics}
\end{figure}

\begin{figure}[htbp]
    \centering
    \includegraphics[width=0.8\linewidth]{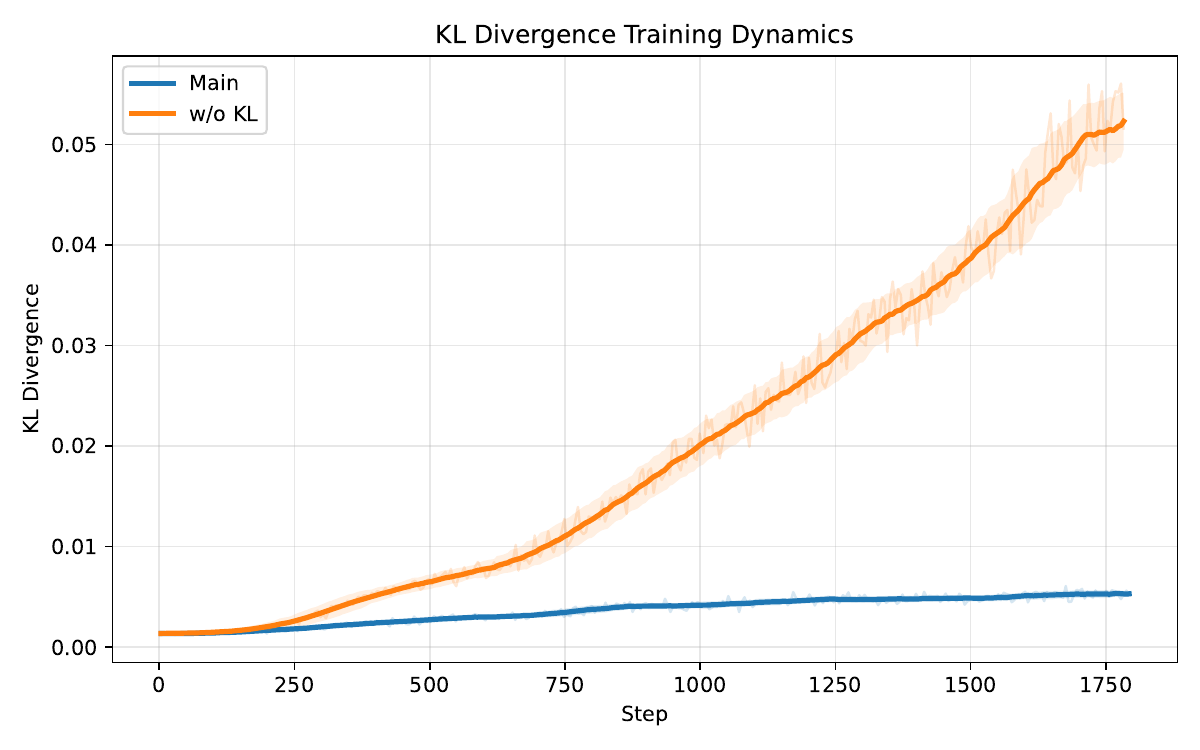}
    \caption{KL divergence dynamics during AeSlides training, compared with \textit{w/o KL Div.} ablation variant. (\textit{smoothed with a rolling window of size 30; mean ± std})}
    \label{fig:kl-dynamics}
\end{figure}

\begin{figure}[htbp]
    \centering
    \includegraphics[width=0.8\linewidth]{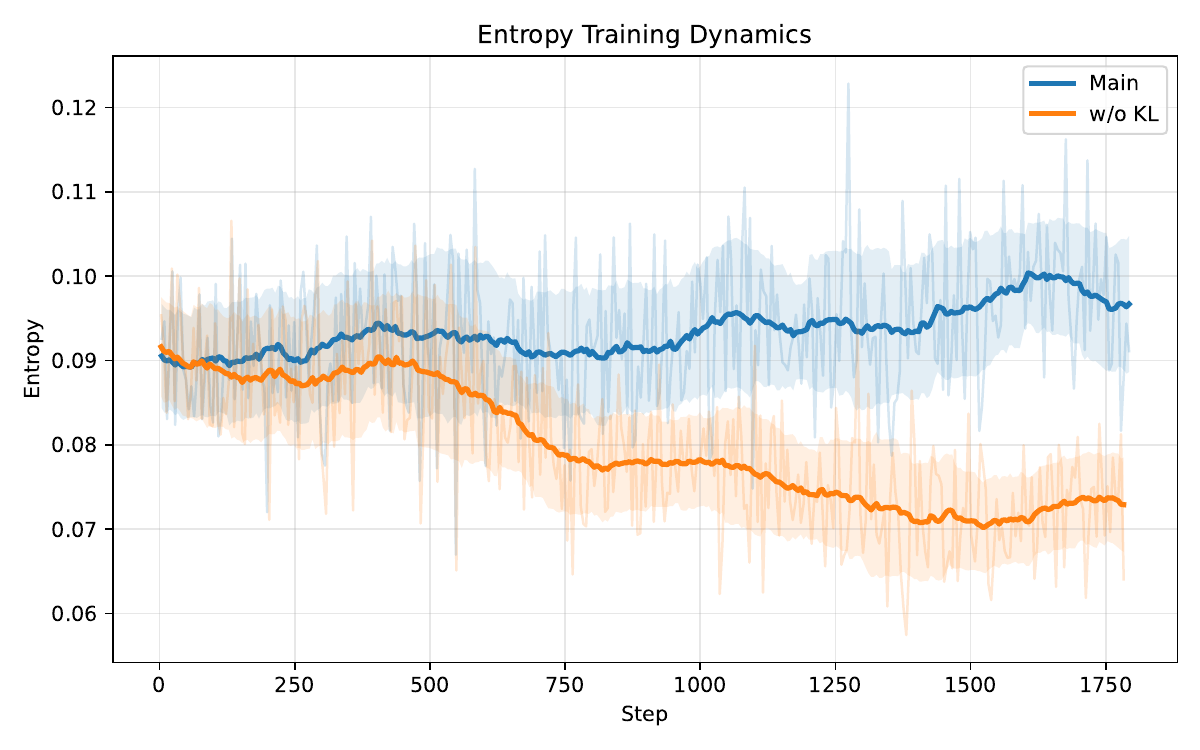}
    \caption{Entropy dynamics during AeSlides training, compared with \textit{w/o KL Div.} ablation variant. (\textit{smoothed with a rolling window of size 30; mean ± std})}
    \label{fig:entropy-dynamics}
\end{figure}

\begin{figure}[htbp]
    \centering
    \includegraphics[width=0.8\linewidth]{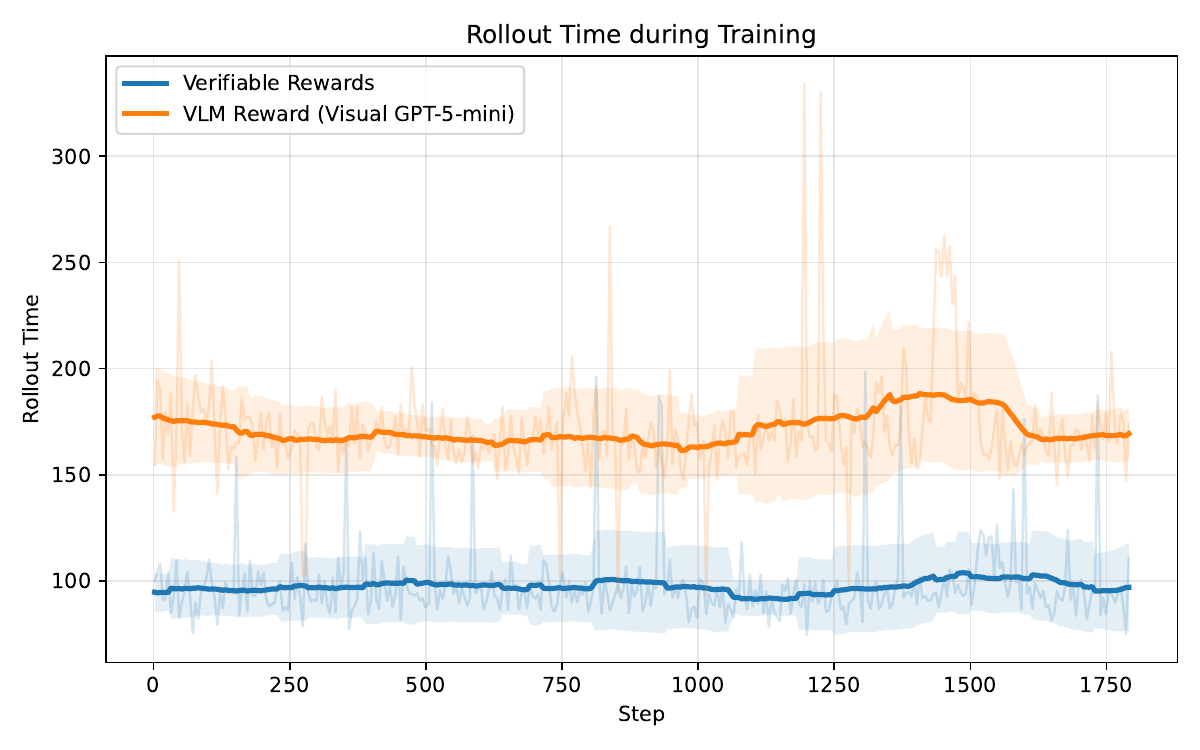}
    \caption{Rollout time during AeSlides training. (\textit{smoothed with a rolling window of size 50; mean ± std})}
    \label{fig:rollout-time-dynamics}
\end{figure}

\subsection{Training Dynamics}
\label{appx:training-dynamics}

We further provide additional visualizations to analyze the training dynamics of AeSlides. As shown in Figure~\ref{fig:reward-dynamics}, AeSlides achieves a steady increase in total reward on both the training and evaluation sets.
Figures~\ref{fig:kl-dynamics} and~\ref{fig:entropy-dynamics} present the evolution of KL divergence and policy entropy during training for both the full method and the \textit{w/o KL Div.} ablation variant. Without the KL constraint, the updates become overly aggressive, leading to a rapid increase in KL divergence. This is accompanied by a collapse of the policy space: the model converges to overly stable and conservative design patterns, discarding the diverse patterns learned during the SFT stage. Although such behavior can yield higher reward, it deviates from the intended objective of generating high-quality slides (see also the case study in~\ref{appx:kl-ablation}).
Figure~\ref{fig:rollout-time-dynamics} compares the rollout time of AeSlides and the VLM-based reward model (Visual GPT-5-mini). The reported time corresponds to the total wall-clock time for rollout generation, including both decoding and reward computation.

\begin{figure*}[htbp]
    \centering
    \includegraphics[width=0.9\linewidth]{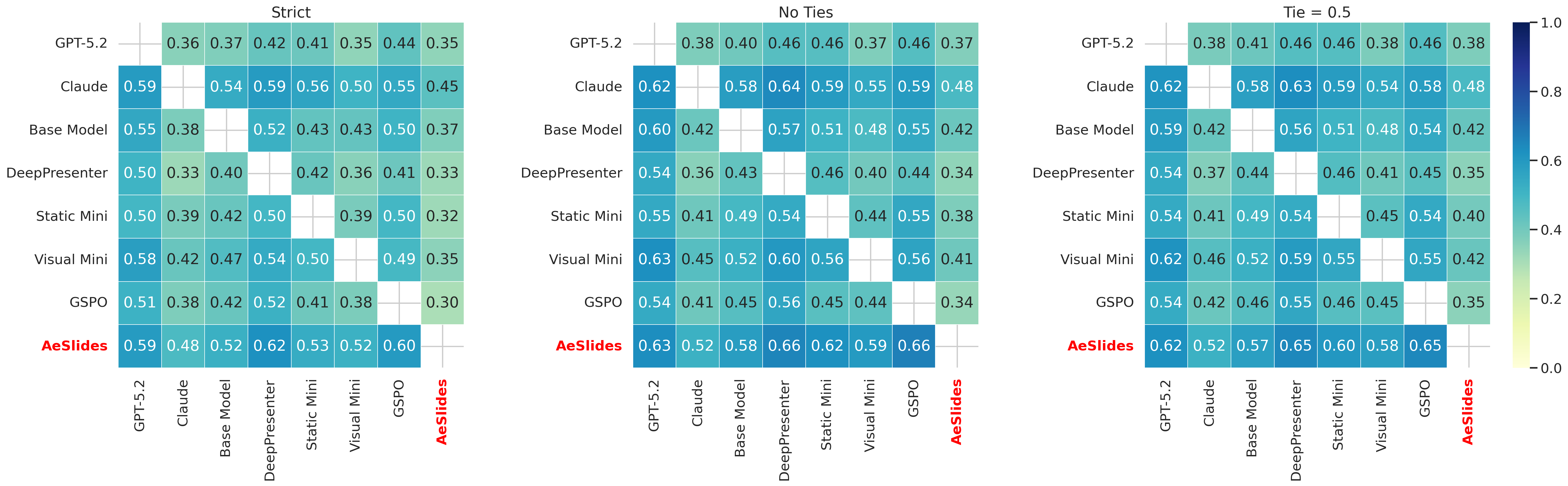}
    \caption{Pairwise win rates of all variants based on human evaluation scores. \textit{Left}: strict win rate; \textit{Center}: win rate that exclude ties; \textit{Right}: Ties contributed as 0.5 wins.}
    \label{fig:pairwise-win-rates}
\end{figure*}

\begin{figure}[htbp]
    \centering
    \includegraphics[width=0.8\linewidth]{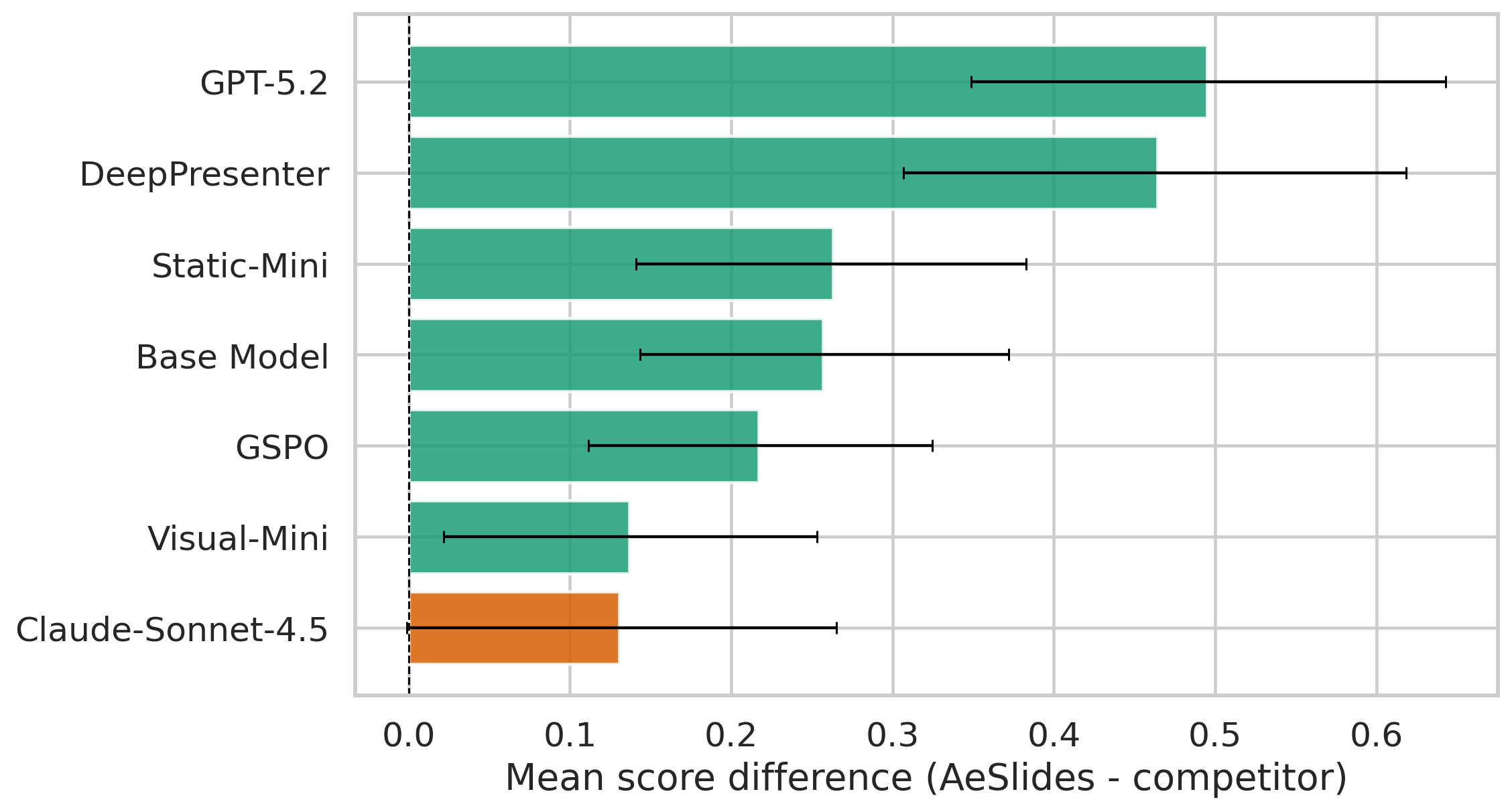}
    \caption{Mean score differences between AeSlides and other variants with bootstrap confidence intervals for statistical significance analysis.}
    \label{fig:significance}
\end{figure}

\begin{figure}[htbp]
    \centering
    \includegraphics[width=0.7\linewidth]{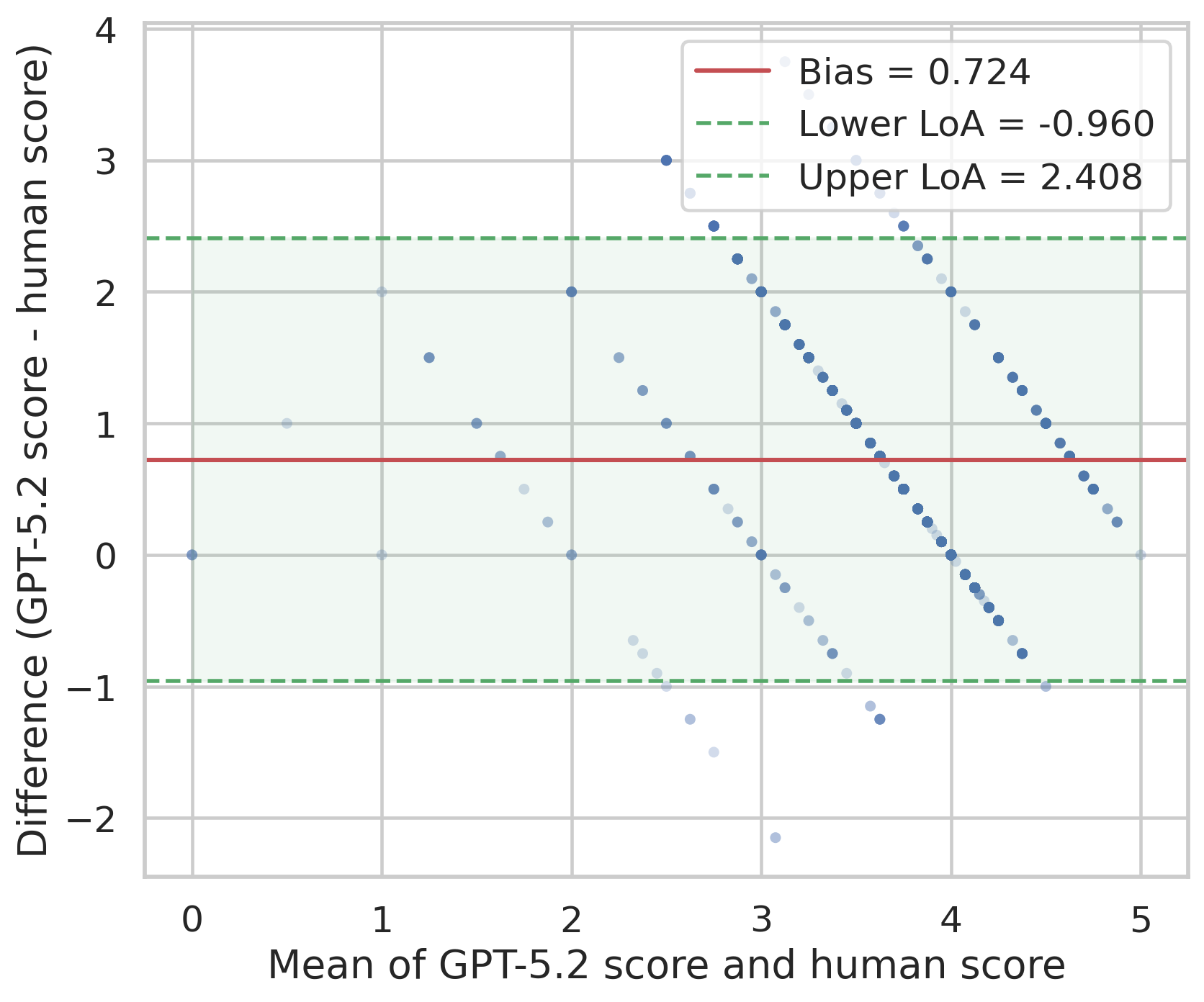}
    \caption{Agreement (Bland-Altman Plot) between human evaluation scores and VLM scores.}
    \label{fig:bland-altman}
\end{figure}

\begin{figure}[htbp]
    \centering
    \includegraphics[width=0.8\linewidth]{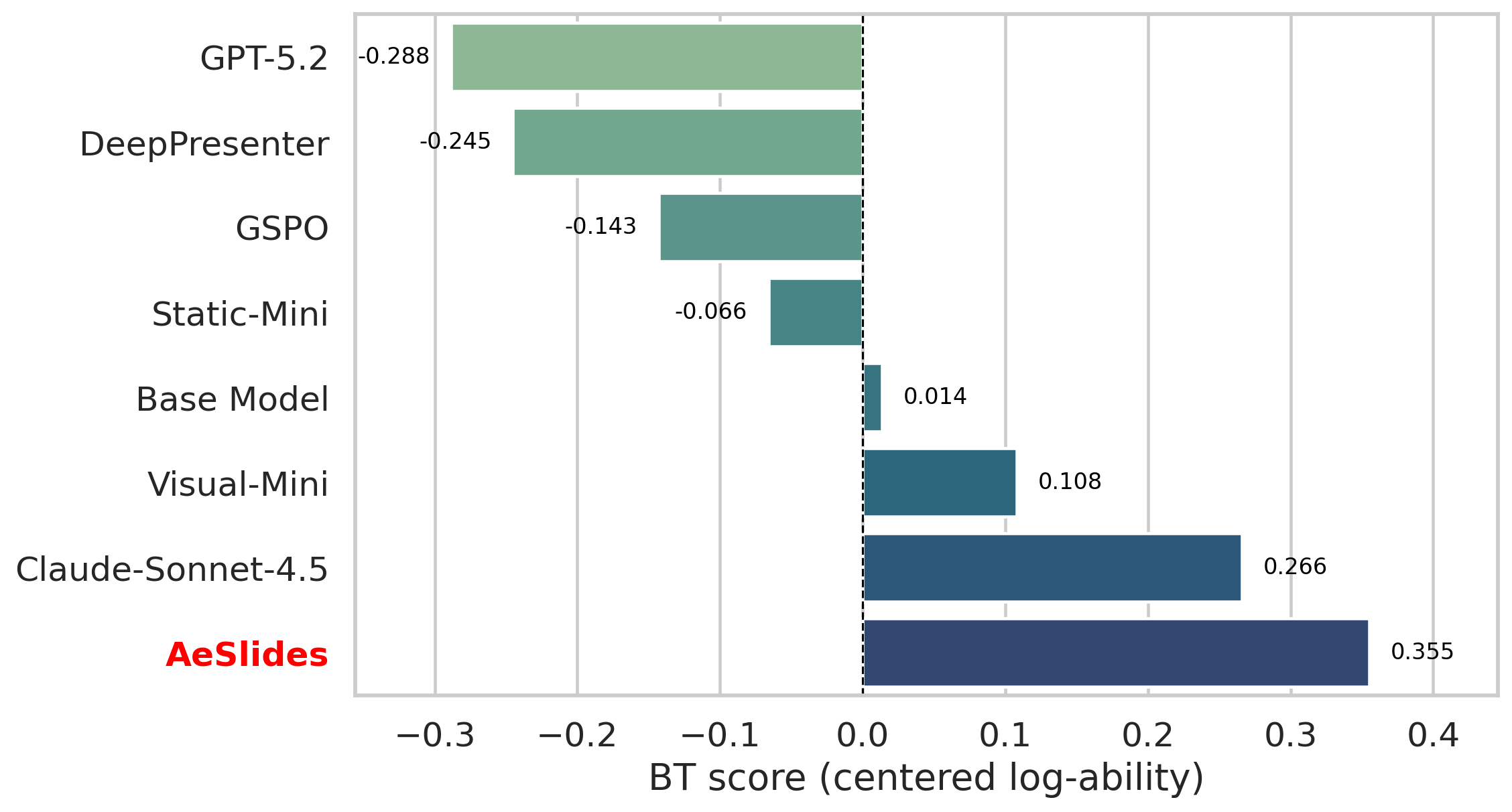}
    \caption{Bradley-Terry model scores of all variants.}
    \label{fig:bradley-terry}
\end{figure}

\subsection{Additional Details of Human Evaluation}
\label{appx:human-eval}

To further conduct a more in-depth analysis of human evaluation results, we compute additional statistical metrics and provide corresponding visualizations, as presented below.

\textbf{Significance of performance improvements with AeSlides.} We conduct a paired, sample-level statistical evaluation to determine whether AeSlides exhibits statistically significant performance improvements over competing variants. Over all evaluation samples, we compute per-sample score differences (AeSlides - competitor). We further estimate the uncertainty via bootstrap-based 95\% confidence intervals, as shown in Figure~\ref{fig:significance}. To assess statistical significance, we perform the Wilcoxon signed-rank test and the paired t-test, with Holm correction applied for multiple comparisons. The Wilcoxon test shows that AeSlides achieves statistically significant improvements over all compared models except Claude-Sonnet-4.5 (Holm-adjusted $p = 0.067$, which is slightly above $0.05$). Consistently, as illustrated in Figure~\ref{fig:significance}, the bootstrap confidence intervals lie entirely above zero for all variants except Claude-Sonnet-4.5, whose interval slightly overlaps zero. The paired t-test, in contrast, indicates statistically significant improvements for all variants. Overall, AeSlides shows consistent improvements across all variants, with statistically significant gains over most baselines under non-parametric testing, and competitive performance compared to the proprietary model Claude-Sonnet-4.5.

\textbf{Agreement between human evaluation and VLM scores.} We plot the Bland-Altman diagram between human evaluation scores and GPT-5.2 scores under a unified scoring protocol, as shown in Figure~\ref{fig:bland-altman}. The plot indicates that GPT-5.2 systematically overestimates scores relative to human judgments, with wide limits of agreement, suggesting weak consistency. In addition, clear heteroscedasticity is observed (i.e., a regression-to-the-mean tendency). Together with the statistical metrics reported in the main paper (Spearman 0.22, QWK 0.19), these results further confirm that current VLM scores are largely unreliable and unsuitable for rewarding and evaluation in slide generation tasks.

\subsection{Additional Experimental Results}
\label{appx:additional-results}

To further illustrate comparative model performance and mitigate potential inter-sample scale inconsistency in human evaluation, we additionally visualize pairwise win rates based on per-sample human evaluation scores, as shown in Figure~\ref{fig:pairwise-win-rates} (under three different definitions of win rate). Furthermore, we fit a {Bradley-Terry} model to the pairwise preference data, yielding the BT scores presented in Figure~\ref{fig:bradley-terry}. Both figures indicate that AeSlides consistently outperforms all variants.


\begin{figure*}[htbp]
    \centering
    \includegraphics[width=0.9\linewidth]{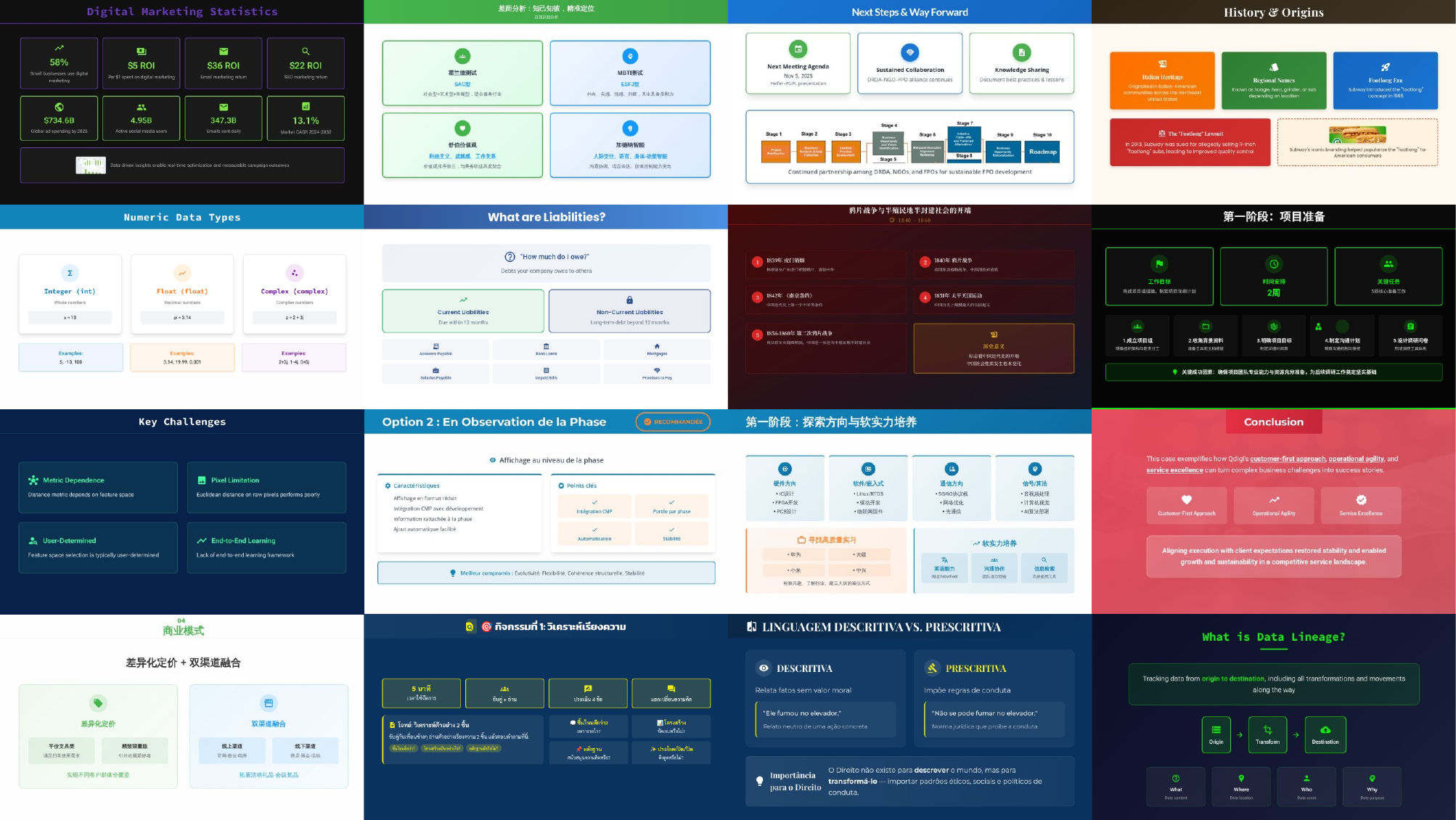}
    \caption{Failure cases of KL ablation. The generated slides collapse into a limited space of conservative design patterns.}
    \label{fig:kl-ablation-case}
\end{figure*}

\subsection{Failure Cases of KL Ablation}
\label{appx:kl-ablation}

We present a series of outputs from the \textit{w/o KL Div.} ablation variant in Figure~\ref{fig:kl-ablation-case}. As observed, removing the KL divergence during reinforcement learning leads the model to adopt a highly templated design pattern and rapidly collapse its policy space, discarding other diverse but error-prone patterns acquired during the SFT stage. This behavior is undesirable. Our objective is for the model to retain previously learned design patterns and perform fine-grained adjustments to HTML attributes so as to better satisfy aesthetic criteria. Based on this observation, we retain the KL divergence regularization term to constrain policy updates.
We also explored alternative training and evaluation strategies. In particular, we examined the Vendi Score~\cite{friedman2023the} used in DeepPresenter~\cite{zheng2026deeppresenter}, which measures diversity via the eigenvalue entropy of feature similarity matrices extracted by DINOv2. However, under our prefix-conditioned setting, this metric fails to provide a reliable estimate of diversity. Instead, it tends to reward behaviors that deviate from user instructions, the style of preceding slides, and the reasoning chain-of-thought, which is undesirable. Therefore, we do not adopt this metric for training or evaluation in this work, and leave the design of more effective diversity metrics for future work.

\begin{figure*}[htbp]
    \centering
    \includegraphics[width=0.9\linewidth]{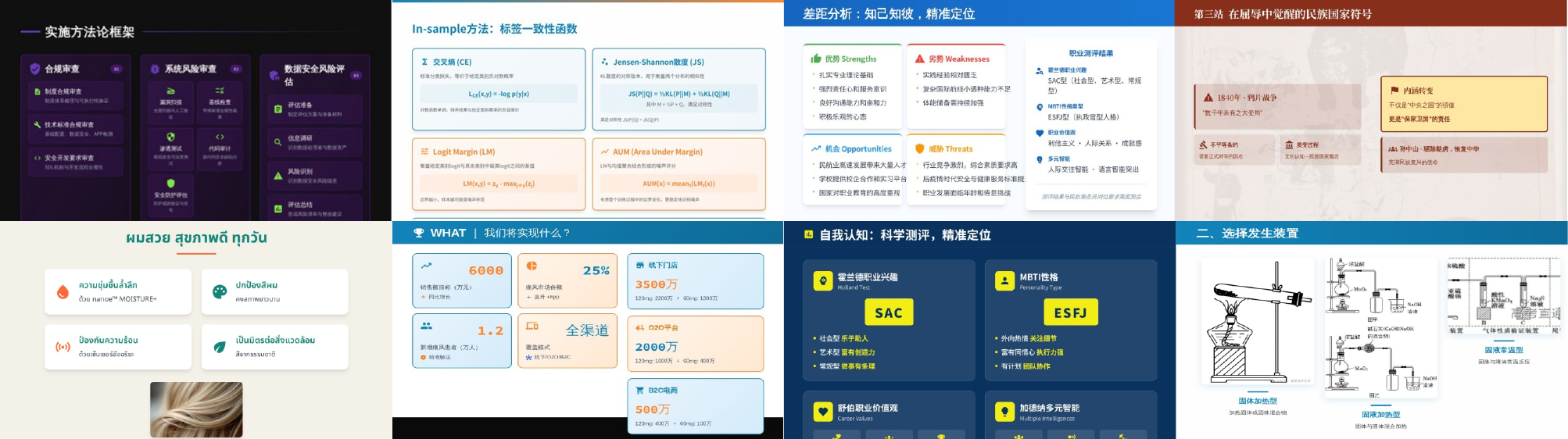}
    \caption{Failure cases of AeSlides.}
    \label{fig:failure-cases}
\end{figure*}

\subsection{Failure Cases of AeSlides}

We also observe that, despite being trained with explicit aesthetic supervision, AeSlides may still exhibit certain failure modes at inference time. Based on feedback from human annotators, we collect a subset of representative failure cases and present them in Figure~\ref{fig:failure-cases}. The main patterns are summarized as follows.
(i) Overly strong conditions specified in the user query or the model-generated plan. For example, when the user explicitly requires displaying all content on each slide while the total content volume is excessive, the model struggles to simultaneously satisfy the instruction and maintain a balanced aspect ratio. Similar issues arise when the content is too sparse, leading to excessive whitespace. We attribute such failures primarily to limitations in the user query itself or the planning stage in the prefix condition, rather than to deficiencies in the model's aesthetic capability.
(ii) Fine-grained aesthetic attributes that are not explicitly supervised, such as typography (e.g., font family and size) and color schemes. These issues occur relatively infrequently in practice and are more difficult to evaluate reliably. Therefore, this work primarily focuses on higher-level layout aesthetics. Nevertheless, AeSlides is an extensible framework: if verifiable metrics for these dimensions become available, they can be seamlessly incorporated into the framework, which we leave as future work.

\begin{figure*}[htbp]
    \centering
    \includegraphics[width=0.9\linewidth]{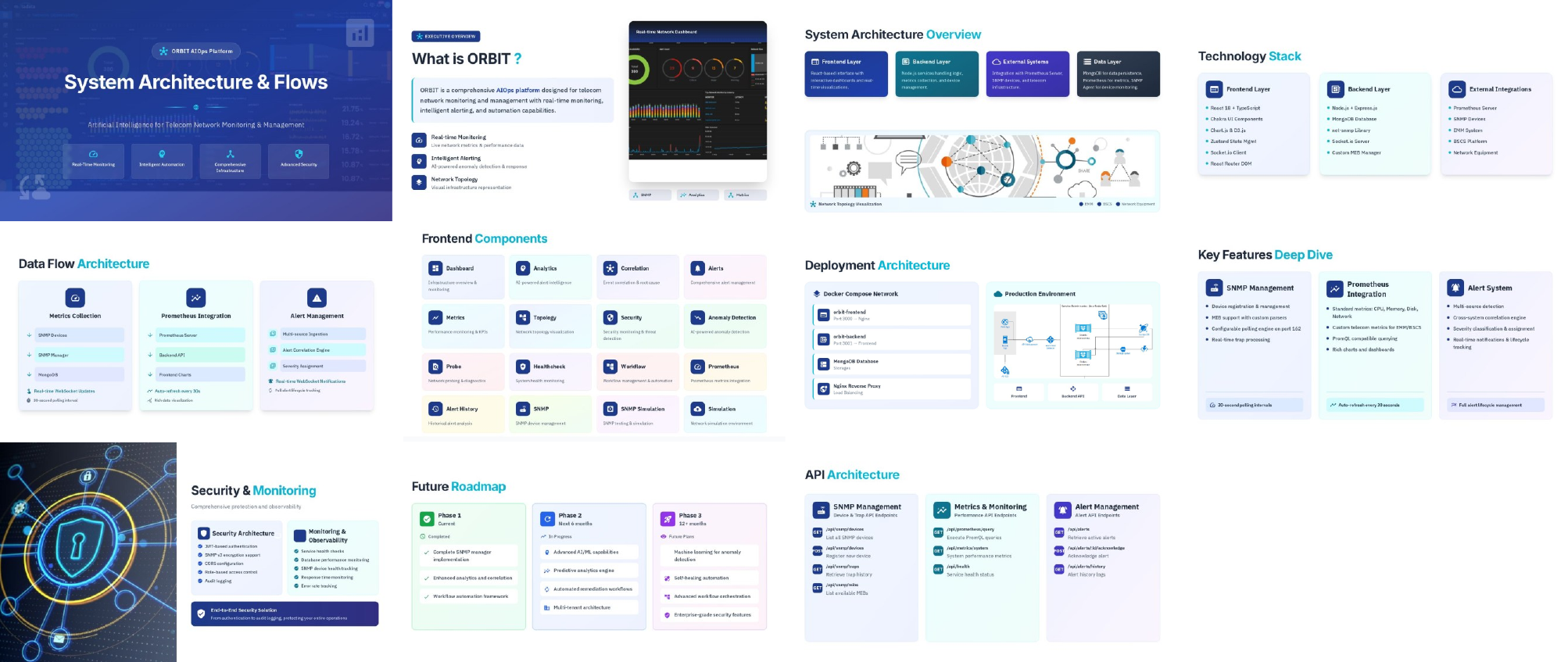}
    \caption{End-to-end generation case of AeSlides (i). (\textit{User query: Create PPT from this: <uploaded document about XXXXX Platform>})}
    \label{fig:e2e-case-1}
\end{figure*}

\begin{figure*}[htbp]
    \centering
    \includegraphics[width=0.9\linewidth]{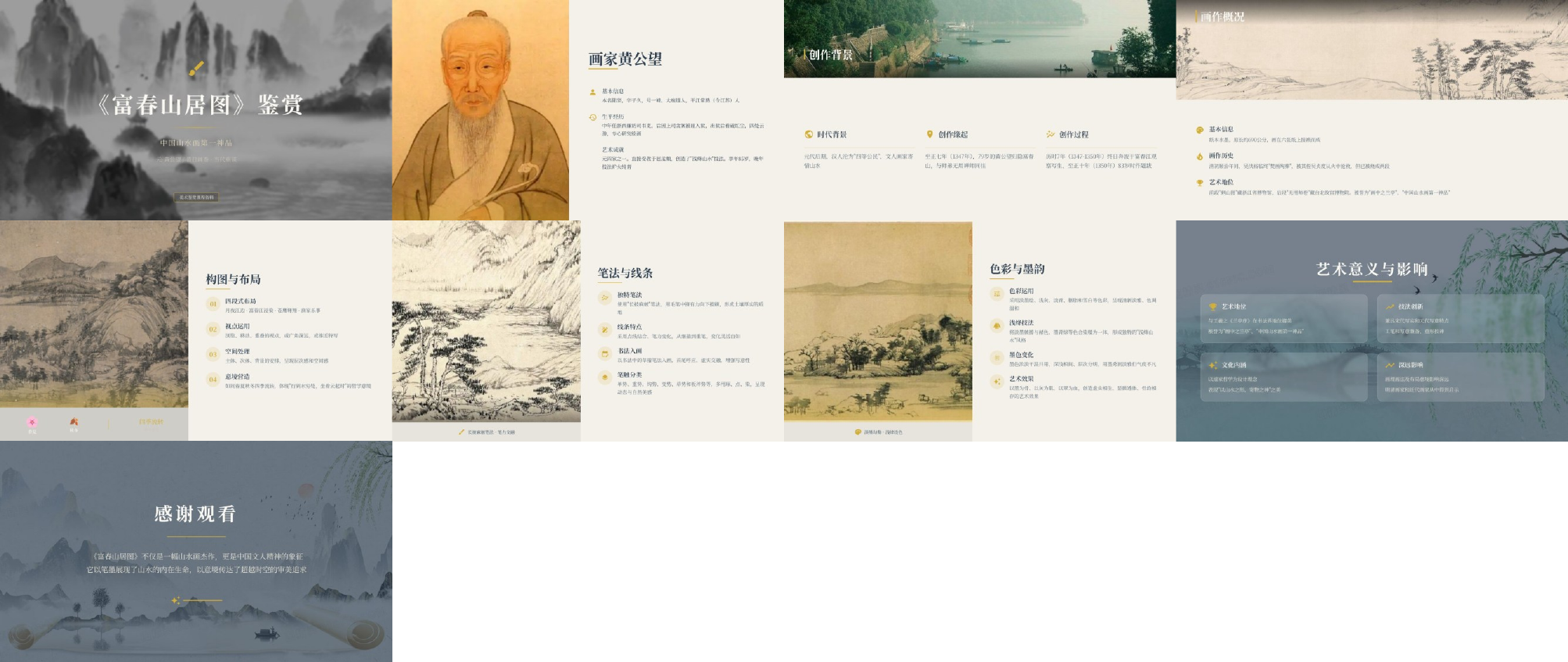}
    \caption{End-to-end generation case of AeSlides (ii). (\textit{User query (Chinese translated into English): Create a PPT to introduce and appreciate the painting "Dwelling in the Fuchun Mountains" for students to appreciate in art class. The content should include: introduction of the artist, background, techniques (composition, color, characters, lines, etc.), significance, etc. The content should be rich and detailed. The PPT layout should be standard size (1280×720 pixels), and the PPT background and style should match the painting.})}
    \label{fig:e2e-case-2}
\end{figure*}

\subsection{End-to-End Case Study of AeSlides}

To further demonstrate that single-page-level aesthetic supervision does not compromise slide-deck-level generation quality, we conduct additional end-to-end generation experiments. Representative cases are shown in Figures~\ref{fig:e2e-case-1} and~\ref{fig:e2e-case-2}. The results indicate that, while achieving improved aesthetic quality, the model maintains strong content quality and consistent cross-slide stylistic coherence.


\FloatBarrier
\newpage

\begin{figure*}[htbp]
\begin{tcolorbox}[colframe=blue!50!black, colback=blue!10, coltitle=white, fonttitle=\bfseries, title=AeSlides-7k Case]
<system>\\

You are GLM-AeSlides, an AI assistant designed to help users create professional slides......\\

<user>\\

Create a slide deck about Endolymphatic hydrops imaging\\

<assistant>\\

<think> The user is asking about "Endolymphatic hydrops imaging." This appears to be a medical imaging topic related to endolymphatic hydrops, which is a condition affecting the inner ear. To provide comprehensive information, I should:\textbackslash n\textbackslash n1. First understand what endolymphatic hydrops is - it\'s typically associated with Ménière\'s disease and involves abnormal....... </think> 

I'll search for information about endolymphatic hydrops imaging techniques......\\

<tool> {\{'name': 'search', 'arguments': \{'queries': ['endolymphatic hydrops imaging techniques', 'MRI for endolymphatic hydrops diagnosis', "imaging modalities for Ménière's disease", 'advances in endolymphatic hydrops imaging']\}\}} </tool> 

Output: 1. Consensus on MR Imaging of Endolymphatic Hydrops in https://pmc.ncbi.nlm.nih.gov/articles/PMC9096015......\\

<think> Based on the search results, I can see that there are several important articles about endolymphatic hydrops imaging. The most recent and comprehensive information appears to be in the consensus document from 2022, which would provide standardized approaches to imaging. There are also articles discussing specific MRI techniques, including both contrast-enhanced and non-contrast methods.\textbackslash n\textbackslash nKey points I've noticed from the search results......</think> 

I'll visit the consensus document on MR imaging of endolymphatic hydrops to get the most comprehensive and up-to-date information on imaging techniques.\\

......\\

<think> I have successfully gathered comprehensive information about endolymphatic hydrops imaging and found some relevant MRI images. Now I should create a presentation that covers:\textbackslash n\textbackslash n1. Introduction to Endolymphatic Hydrops and Meniere\'s Disease\textbackslash n2. Imaging Techniques:\textbackslash n   - Contrast-enhanced MRI (3D-FLAIR and 3D-real IR sequences)\textbackslash n   - Contrast administration methods (intratympanic vs. intravenous)\textbackslash n   - Non-contrast MRI techniques\textbackslash n3. Grading Systems (Baráth and Bernaerts)\textbackslash n4. Clinical Applications and Diagnostic Accuracy\textbackslash n5. Recent Advances and Future Directions\textbackslash n\textbackslash nI should plan the slides to be comprehensive yet concise, with a focus on the key imaging techniques, grading systems, and clinical applications. The presentation...... </think>\\

<tool> {\{'name': 'initialize\_design', 'arguments': \{'title': 'Endolymphatic Hydrops Imaging', 'slide\_name': 'endolymphatic\_hydrops\_imaging'\}\}} </tool>\\

<think> The presentation has been successfully initialized. Now I need to create each of the 9 slides according to the plan...... </think> I'll now create the first slide - the title slide for the presentation.\\

<tool> {\{'name': 'insert\_page', 'arguments': \{'index': 1, 'html': '<!DOCTYPE html>\textbackslash n<html lang="en">\textbackslash n<head>\textbackslash n<meta charset="utf-8"/>\textbackslash n......'\}\}} </tool> \\

\textcolor{red}{<think> Great! The title slide has been created successfully. Now I need to create the second slide about the introduction to endolymphatic hydrops and Meniere's disease. </think> Now I'll create the second slide introducing endolymphatic hydrops and Meniere's disease.}\\ 

\textcolor{red}{<tool> {\{'name': 'insert\_page', 'arguments': \{'index': 2, 'html': '<!DOCTYPE html>\textbackslash n<html lang="en">\textbackslash n<head>\textbackslash n<meta charset="utf-8"/>\textbackslash n......'\}\}} </tool> \\}

\end{tcolorbox}
\caption{Case of prefix-conditioned generation prompt in AeSlides-7k. The \textcolor{red}{Red} part indicates the rollout generation. For readability, we apply necessary simplifications to the original chat template.}
  \label{fig:case-7k}
\end{figure*}

\begin{figure*}[htbp]
\begin{tcolorbox}[colframe=blue!50!black, colback=blue!10, coltitle=white, fonttitle=\bfseries, title=Slide Layout Aesthetic Evaluation Prompt]
You are a Visual QA Assistant specializing in Slide Layout.\\
Your task is to evaluate the overall layout aesthetics of a slide screenshot.\\

Core Evaluation Dimensions (must be jointly considered):\\
1. distorted\_aspect\_ratio:\\
The slide should follow a 16:9 ratio (1280x720). Treat width/height ratios within +/-5\% of 16:9
(roughly 1.69 to 1.87) as acceptable. Deviations or visual distortion should be considered a flaw.\\

2. excessive\_whitespace:\\
Identify empty regions without semantic content.\\
Note: Background fills (gradients, textures) do NOT count as content.\\

3. element\_collision:\\
Check for overlap, occlusion, overflow, clipping, or boundary violations between elements.\\

4. visual\_imbalance:\\
Evaluate whether the layout is visually balanced or biased toward one side.\\

Evaluation Criteria:\\
Judge the slide holistically as a visual composition. Consider layout balance, alignment, spacing, hierarchy, readability, proportional sizing, whitespace distribution, and visual grouping.\\

Strongly penalize:\\
- Incorrect or visually distorted aspect ratio\\
- Large meaningless empty regions (even if visually filled)\\
- Element collisions (overlap, occlusion, overflow, clipping)\\
- Strong visual imbalance (content heavily biased to one side)\\
- Other obvious layout flaws that make the slide look broken, chaotic, empty, or unprofessional\\

Do NOT penalize:\\
- Intentional minimalism or clean negative space when it supports clarity\\
- Minor alignment imperfections that do not affect perception\\

Scoring Rubric (0--5):\\
- 0: Broken layout. Severe issues in $\geq 2$ core dimensions. Unusable.\\
- 1: Major failure. At least one severe issue OR multiple clear issues.\\
- 2: Weak layout. Noticeable issues in $\geq 2$ dimensions. Feels unpolished.\\
- 3: Mixed quality. One clear issue OR multiple mild issues. Usable but lacks refinement.\\
- 4: Strong layout. Most dimensions well-controlled with minor imperfections.\\
- 5: Excellent layout. All four dimensions well-balanced. Cohesive and presentation-ready.\\

Output only the final answer between <ans> and </ans>. \\
For example: <ans>3</ans>\\

Input:\\
\texttt{\{Slide Screenshot\}}
\end{tcolorbox}
\caption{Prompt for VLM-based slide aesthetic overall evaluation.}
\label{fig:vlm-overall-prompt}
\end{figure*}

\begin{figure*}[htbp]
\begin{tcolorbox}[colframe=blue!50!black, colback=blue!10, coltitle=white, fonttitle=\bfseries, title=Slide Issue Detection Prompt]
You are a Visual QA Assistant specializing in slide layout defect detection.\\
Your task is to judge whether a slide screenshot has any obvious layout problems in exactly these categories:\\

1. has\_distorted\_aspect\_ratio:\\
The standard slide aspect ratio is 16:9 (1280x720).\\
Treat width/height ratios within +/-5\% of 16:9
(roughly 1.69 to 1.87) as acceptable. Outside that tolerance, or when the slide visibly appears stretched,
squashed, or improperly scaled, it should be treated as a significant aspect-ratio defect.\\

2. has\_excessive\_whitespace:\\
There are abnormally large regions without semantic content.\\
Note: Background fills (gradients, textures, color blocks) DO NOT count as content.\\

3. has\_element\_collision:\\
Elements visibly overlap, occlude each other, overflow containers, or extend beyond boundaries.\\
Do NOT flag minor or visually harmless overlays (e.g., subtle badges, light decorations).\\

4. has\_visual\_imbalance:\\
The visual center of gravity is clearly biased to one side (left/right/top/bottom), making the layout feel unstable or uncomfortable.\\

Judging Rules:\\
- Only flag obvious and visually significant issues\\
- Do NOT flag minor imperfections\\
- Do NOT penalize intentional clean layouts unless clearly problematic\\

Output the final answer between <ans> and </ans>. Return a JSON object with exactly these keys:\\
\{\\
\begin{tabular}{l l}
"has\_distorted\_aspect\_ratio": & true | false, \\
"has\_excessive\_whitespace":   & true | false, \\
"has\_element\_collision":      & true | false, \\
"has\_visual\_imbalance":       & true | false
\end{tabular}\\
\}\\

For example:\\
<ans>\{\\
\begin{tabular}{l l}
"has\_distorted\_aspect\_ratio": & false, \\
"has\_excessive\_whitespace":   & true, \\
"has\_element\_collision":      & false, \\
"has\_visual\_imbalance":       & true 
\end{tabular}\\
\}</ans>\\

Input:\\
\texttt{\{Slide Screenshot\}}\\

\end{tcolorbox}
\caption{Prompt for VLM-based slide aesthetic issue detection.}
  \label{fig:vlm-prompt}
\end{figure*}

\end{document}